\documentclass[letterpaper]{article} % DO NOT CHANGE THIS
\usepackage[preprint]{aaai2027}  % arXiv preprint: shows authors, drops the AAAI copyright slug
\usepackage[hyphens]{url}  % DO NOT CHANGE THIS
\usepackage{graphicx} % DO NOT CHANGE THIS
\usepackage{natbib}  % DO NOT CHANGE THIS AND DO NOT ADD ANY OPTIONS TO IT
\usepackage{caption} % DO NOT CHANGE THIS AND DO NOT ADD ANY OPTIONS TO IT
\usepackage{algorithm}
\usepackage{algorithmic}
\usepackage{newfloat}
\usepackage{listings}
\DeclareCaptionStyle{ruled}{labelfont=normalfont,labelsep=colon,strut=off}
\floatstyle{ruled} 
\newfloat{listing}{tb}{lst}{}
\floatname{listing}{Listing}
\usepackage{booktabs}
\usepackage{multirow}
\usepackage{amsmath}
\usepackage{amssymb}
\usepackage{xspace}
\usepackage{placeins}  % \FloatBarrier (AAAI-safe, no hyperref)
\usepackage{needspace}  % \needspace keeps a heading with its first lines (no vspace hack)
\usepackage{microtype}  % protrusion+expansion: tighter line-breaking, fewer widows
\newcommand{\thedit}{\theta_{\mathrm{edit}}}
\newcommand{\thgoal}{\theta_{\mathrm{Goal}}}
\newcommand{\thbase}{\theta_{\mathrm{base}}}
\newcommand{\tauv}{\boldsymbol{\tau}}
\newcommand{\cmark}{\checkmark}
\newcommand{\xmark}{$\times$}

\newcommand{\piZeroFive}{$\pi_{0.5}$\xspace}

\title{Suppression Sticks, Locality Is Fragile:\\
A Closed-Loop Target-and-Control Audit of Task-Vector Negation in VLA Policies}

\author{
    Shaoguang Wang\textsuperscript{\rm 1},
    Weiyu Guo\textsuperscript{\rm 1}\corresponding{},
    Rushi Dai\textsuperscript{\rm 1},
    Yiren Zhao\textsuperscript{\rm 1},
    Yandong Guo\textsuperscript{\rm 2},
    Hui Xiong\textsuperscript{\rm 1,\rm 3}\corresponding{}
}
\affiliations{
    \textsuperscript{\rm 1}The Hong Kong University of Science and Technology (Guangzhou)\\
    \textsuperscript{\rm 2}AI\textsuperscript{2} Robotics\\
    \textsuperscript{\rm 3}The Hong Kong University of Science and Technology
}
\newif\ifmain
\mainfalse     %%% <== MAIN paper. Change to \mainfalse for the full build.  \maintrue
\newcommand{\GoalAllOffdiag}{37.7\%}   % all-10-row raw off-diagonal mean
\newcommand{\GoalSepCRnorm}{57.8\%}     % 5 separation rows, baseline-normalized CR mean (recompute_ninecontrol_stats.py; per-row 43.9,64.3,51.5,65.4,63.9)
\newcommand{\GoalSepCRrange}{\text{44--65\%}}  % per-row normalized CR range; \text{} forces text-mode en-dash even inside $...$ (avoids math double-minus rendering)
\newcommand{\GoalTZeroCtrl}{43.9\%}     % t0 row control mean (9 controls)
\newcommand{\SpatialOff}{61.7\%}        % spatial off-diag mean (inflated by resistant rows)

\newcommand{\HeldoutCRnorm}{52\%}          % mean baseline-normalized control retention across 5 sep rows (9-control, consistent)
\begin{document}
\maketitle

%%=============================================================
\begin{abstract}
Task-vector arithmetic offers a closed-form way to modify a model, yet its behavioral locality remains unclear in closed-loop robot control. We present a target-and-control audit of per-skill task-vector subtraction from multitask vision-language-action (VLA) policies. Across all ten LIBERO-Goal skills, subtraction produces three qualitatively different regimes: target--control separation for five skills, resistance for three, and global collapse for two. On held-out initial states, the five suppressible targets remain at 0\% success; however, mean baseline-normalized control retention is only 52\%, and each target-suppressing edit materially harms at least one nominally unrelated control. Additional Goal panels show separation across tested policies with continuous-regression, discrete-token, and flow-matching action heads, whereas we observe no clean separation on Spatial and control collapse on the tested Object and Long-horizon panels. Mean task-vector cosine does not account for this variation. A matched-norm control identifies a local sign asymmetry around one Goal anchor, while multi-vector outcomes vary with anchor and scale. Retain-aware gradient baselines provide data-dependent comparators but require removal-time data and optimization; subtraction is data- and gradient-free only at edit time, assuming precomputed expert deltas. Finally, a single-skill relearning probe is consistent with behavioral masking, not certified unlearning. These results characterize task-vector subtraction as a fast but brittle intervention and underscore the need for closed-loop target-and-control evaluation when assessing locality in embodied model editing.
\end{abstract}

%%=============================================================
\section{Introduction}

\begin{figure*}[t]
\centering
\includegraphics[width=\textwidth]{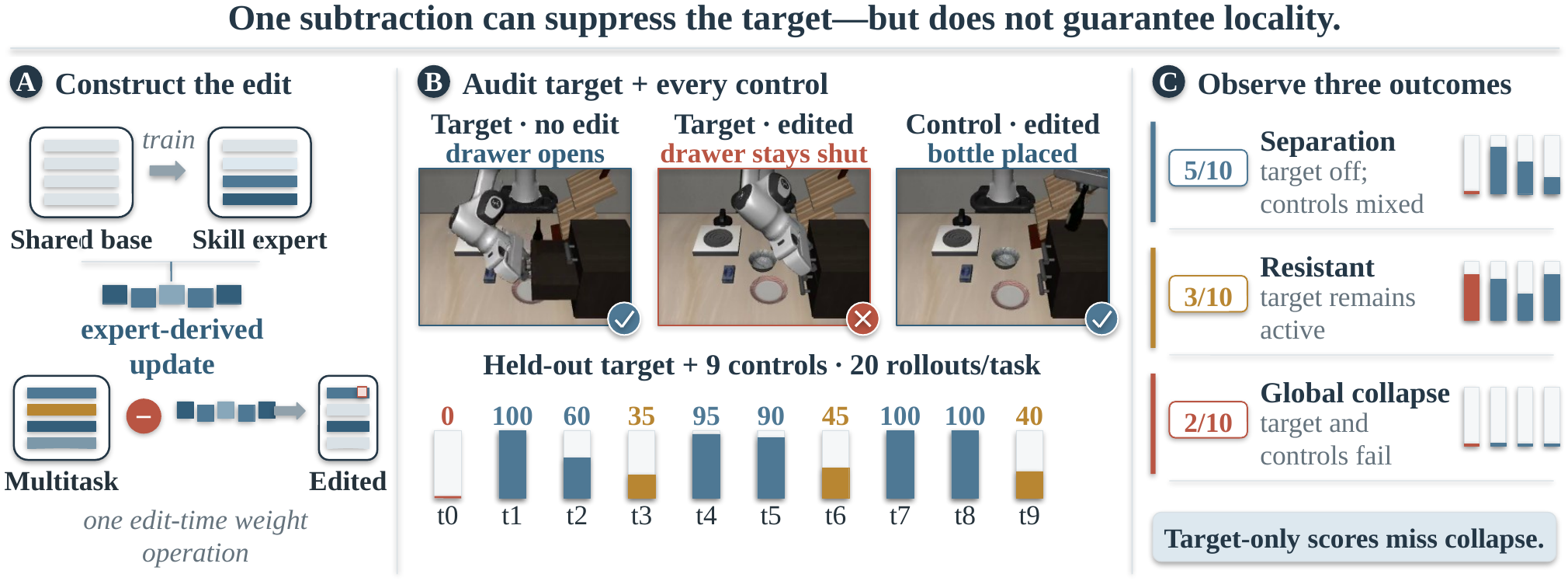}
\caption{\textbf{A single edit-time subtraction removes the target skill, but its locality is fragile and heterogeneous.} We derive an
expert update from a single-skill expert and subtract it from a multitask VLA in
one edit-time operation, then roll out the target and every control behavior. All
frames are real MergeVLA LIBERO-Goal rollouts at $\alpha{=}0.75$: the $t_0$
target opens the drawer without the edit and fails after subtraction (same
initial state), while a control skill still succeeds under the same edited
policy (a deterministic lowest-index successful rollout, illustrative). The
held-out strip gives the edited success rate of the target and all nine controls
for this edit (target off; controls mixed). Across ten LIBERO-Goal skills the
edit yields three regimes: five target--control separation rows, three resistant
targets, and two global collapses. Formal notation is in the Setup section.}
\label{fig:teaser}
\end{figure*}

As multitask Vision-Language-Action (VLA) policies are deployed, practitioners
increasingly need to \emph{remove} a single learned behavior (one that is
unsafe, deprecated, or requested for deletion) without retraining the whole
policy or degrading the skills that remain. Task arithmetic~\cite{ilharco2023editing}
offers a tempting shortcut: a per-skill task vector
$\tauv_i=\theta_{\mathrm{exp},i}-\thbase$ (Eq.~\ref{eq:skillvec}) subtracted from
a multitask checkpoint, $\thedit=\thgoal-\alpha\,\tauv_i$, is a closed-form edit
that, given precomputed experts, needs no removal-time data or gradients and
runs in under a second on CPU.

But whether such an edit is \emph{local}, suppressing the target while leaving
the other skills intact, has only been checked where the edited object is a
static label. In closed-loop robot control the edited object is a
\emph{behavior}, and locality must be judged by rolling out both the target and
every other skill after the edit. A weight perturbation that looks benign under
a single forward pass can still derail an unrelated skill over a full rollout,
so the safety-relevant question (does removing one skill silently break
another?) cannot be inferred from static-loss diagnostics.

Prior work leaves this gap open. Task-arithmetic studies treat \emph{addition}
as the headline operation (merging, model soups~\cite{wortsman2022soups,ilharco2023editing,yadav2023ties,yu2024dare})
and report negation only as a forgetting side-effect on static
models~\cite{ilharco2023editing}. Machine-unlearning methods that do target
removal need the retain set and training-scale
compute~\cite{bourtoule2021sisa,golatkar2020eternal,guo2020certified}, the only
data-free approach still backpropagates~\cite{chundawat2023zero}, and the closest
VLA removal work~\cite{ranjan2026vlaf} is gradient-based and data-dependent at
removal time. No prior study audits a closed-form subtraction through
closed-loop target-and-control robot behavior.

We therefore design a \emph{closed-loop audit} of per-skill task-vector
subtraction on VLA policies~\cite{fu2025mergevla,kim2024openvla}
(Fig.~\ref{fig:teaser}): for each skill we subtract its vector from the goal
checkpoint $\thgoal$, then measure closed-loop success of the target
\emph{and} all other skills, across all ten LIBERO-Goal skills, four suites,
and three action-head families, plus matched sign and gradient-baseline probes.

The audit yields a clear but cautionary picture. Subtraction produces three
regimes: for \emph{five} skills a \textbf{target--control separation}
(target $\leq\!15\%$, in-sample control retention $\GoalSepCRnorm$); \emph{three} skills
\textbf{resist} suppression; and \emph{two} suppress the target but collapse
every control (\textbf{global collapse}). Even among the separation rows,
locality is fragile: on held-out states mean control retention falls to $\HeldoutCRnorm$, ranging from $78\%$ down to $4\%$, and most edits zero at least one nominally unrelated control. The separation pattern recurs across continuous,
discrete-token, and flow-matching action heads but breaks on suites with shared
motor primitives and under multi-skill composition. Mean weight-space cosine does
not predict which controls survive. We thus read single-vector subtraction as
\emph{robust target suppression with fragile, heterogeneous locality}, and argue
that closed-loop target-and-control evaluation, not static loss, is the right
lens for editing embodied policies.

\paragraph{Contributions.} We present the first closed-loop \emph{target-and-control} audit of task-vector subtraction in VLA policies, and find it robustly removes the target skill while remaining brittle in its locality.
\textbf{(1) A three-regime map of subtraction.} On LIBERO-Goal, closed-loop
subtraction splits the ten skills into three regimes (\emph{five} separate target
from control, \emph{three} resist, and \emph{two} collapse every
control), structure that static-label editing cannot see. The separation pattern
recurs across continuous, discrete-token, and flow-matching action heads, and we
document the goal-anchored asymmetry of positive re-addition (\S\,Experiments).
\textbf{(2) Selectivity and its boundary.} We quantify
substantial collateral damage and a task-family boundary: on suites with shared
motor primitives or subgoals, controls collapse with the target
(\S\,Generality and Boundaries). Critically, a held-out full target$\times$all-controls matrix shows target
suppression still generalizes (all five suppressible skills stay at $0\%$) while
\emph{selective control preservation is heterogeneous}: mean control retention
$\HeldoutCRnorm$, spanning $78\%$ down to full collapse across targets.
\textbf{(3) Diagnostics and permanence.} We study the anchor/sign dependence,
representations, loss landscape, cost, and relearning, and show the edit is
consistent with rapidly reversible behavioral masking; a
goal-calibrated, data-dependent, forward-only alignment diagnostic ranks
suppressibility \emph{within} a suite but does not reliably transfer across
suites (Supp.\ Sec.~D and~E, \S\,Discussion and Limitations).

%%=============================================================
\section{Related Work}
\label{sec:related}

\paragraph{Machine Unlearning.}
Machine unlearning~\cite{bourtoule2021sisa,yao2024llm} removes knowledge without full retraining. Exact and certified methods~\cite{bourtoule2021sisa,ginart2019making,guo2020certified,golatkar2020eternal,eldan2023harry} need the retain set and training-scale compute, infeasible for 500M+ VLA models. Gradient-based methods dominate: gradient ascent~\cite{graves2021amnesiac} can cause catastrophic forgetting~\cite{maini2024tofu}, and retain-term or impair--repair variants such as NegGrad+, SCRUB~\cite{kurmanji2023scrub}, UNSIR~\cite{tarun2023fast}, Gradient-Difference~\cite{liu2022continual}, SalUn~\cite{fan2024salun}, NPO~\cite{zhang2024npo}, and zero-shot unlearning~\cite{chundawat2023zero} all use gradients (only the last is data-free). We benchmark against NegGrad, NegGrad+, and Gradient-Difference. TOFU~\cite{maini2024tofu} and MUSE~\cite{shi2024muse} formalize the forget/retain trade-off, and WMDP/RMU~\cite{li2024wmdp} target hazardous capabilities, but all score token likelihood, not behavioral rollouts. Training-free weight surgery~\cite{li2025skill,kim2025knowvec,meng2022rome,meng2023memit,arditi2024refusal} targets language or vision models, not policies. We study a single subtraction (Ilharco's) that is data-free and gradient-free at edit time given precomputed task vectors, characterizing when it suppresses a VLA skill, how controls respond, and where the effect fails (Supp.\ Table~\ref{tab:comparison}).

\paragraph{Task Arithmetic and Model Merging.}
\citet{ilharco2023editing} introduced task vectors $\tau_i = \theta_{\mathrm{ft},i} - \theta_{\mathrm{pre}}$, where negation sharply reduces target-task accuracy while leaving control tasks largely intact for image classifiers. Addition underlies merging~\cite{wortsman2022soups,matena2022fisher,jin2023regmean,yadav2023ties,yu2024dare,yang2024adamerging}; explanations invoke tangent-space disentanglement~\cite{ortizjimenez2023tangent} and generalization analysis~\cite{li2025taskvecprov}. On unlearning, NegMerge~\cite{kim2025negmerge} and CATA~\cite{lin2026cata} apply arithmetic to VLMs. Such work studies negation only on static classification or generation, and has not evaluated task-vector subtraction through closed-loop robot success, where at matched norm a single vector's sign decides selective suppression versus retention, while large goal-anchored re-addition of multiple skill deltas collapses the whole policy (\S\,Single-Vector Sign vs.\ Composition), a divergence invisible in the single-forward-pass setting.

\paragraph{VLA Architectures, Benchmarks, and Unlearning.}
Generalist policies~\cite{brohan2023rt2,octo2024,kim2024openvla,kim2025openvlaoft,black2024pi0,qu2025spatialvla,guo2026brain,li2026spatialmem} map vision and language to actions through a backbone~\cite{karamcheti2024prismatic} and action head, adapted with LoRA~\cite{hu2022lora}. As these policies scale, cheaply adapting or correcting them (without retraining or large data) grows in importance, and a post-hoc weight edit is one such lightweight intervention. Our edit acts on the \emph{parameter} side; a complementary and orthogonal axis of efficiency operates on the \emph{input} side, shrinking the visual-token budget that a multimodal backbone must process through keyframe search and selection for long-video understanding~\cite{guo2025logicframes,wang2026focus} and adaptive frame pruning~\cite{wang2026afp}. These input-side methods and our weight-side intervention target different bottlenecks and could in principle be combined. MergeVLA~\cite{fu2025mergevla} adds skills via LoRA merging (reports 98\% no-edit SR on LIBERO-Spatial); we use their checkpoints as our goal model. \citet{liu2026vlaforget_resist} report that VLAs resist forgetting, which makes on-demand removal both harder and more valuable. Recent VLA continual-learning studies examine retention under sequential adaptation or mitigate forgetting through phase-aware and recovery-oriented experience replay~\cite{liu2026vlaforget_resist,chen2026phaser,karli2026recall}. Our setting is complementary: given an already trained multitask policy, we ask whether a closed-form, post-hoc expert delta can suppress one behavior without replay or removal-time optimization. LIBERO~\cite{liu2023libero} provides Goal, Spatial, Object, and Long10; we evaluate on LIBERO-Goal, with Spatial, Object, and Long10 as cross-suite tests (\S\,Generality and Boundaries). VLA-Forget~\cite{ranjan2026vlaf} is the closest data-dependent VLA removal method; it uses staged, module-selective optimization with forget, retain, and mismatch data, whereas we study a closed-form edit and its closed-loop locality. Our subtraction (data-free and gradient-free at edit time given precomputed task vectors) instead directly characterizes when a single weight-space edit suppresses a skill, how the unrelated control skills respond, and where the effect ultimately fails.
\section{Setup: Task Vectors and Negation in VLAs}
\label{sec:method}

\subsection{Preliminaries}
We build on the MergeVLA framework~\cite{fu2025mergevla}: a multitask goal
model $\thgoal$ trained on $N$ skills, and per-task experts
$\theta_{\mathrm{exp},i}$ fine-tuned from a shared initialization $\thbase$
(pre-LoRA weights) via LoRA. Each model has three components: VLM backbone $W$,
action head AH, and proprio projector PP. Following~\citet{ilharco2023editing}, we define the per-skill \emph{task vector} $\tauv_i$ as the weight delta:
\begin{align}
\tauv_i^W &= W_{\mathrm{exp},i} - W_\text{base},\quad
\tauv_i^{\mathrm{AH}} = \mathrm{AH}_{\mathrm{exp},i} - \mathrm{AH}_\text{init},\notag\\
\tauv_i^{\mathrm{PP}} &= \mathrm{PP}_{\mathrm{exp},i} - \mathrm{PP}_\text{init}.
\label{eq:skillvec}
\end{align}

\subsection{Negation Formula}
The edited model subtracts the scaled skill vector across all three components:
\begin{equation}
\thedit = \thgoal - \alpha \cdot \tauv_i, \quad \alpha > 0.
\label{eq:negation}
\end{equation}
The scalar $\alpha > 0$ controls suppression strength; a sweep selects a
task-specific constrained operating point $\alpha^\star$ under a fixed rule
(smallest $\alpha$ with target SR $=0\%$, then highest mean control SR;
\S\,Experiments). The edit itself is data-free and gradient-free at edit
time (full method in Supp.\ Listing~\ref{lst:core}); the
suppressibility probe (Supp.\ Sec.~E) is a \emph{data-dependent, gradient-free,
forward-only} diagnostic: it \emph{does} read a small sample of skill-specific
inputs, so it is not data-free.

\subsection{Multi-Skill Negation}
To simultaneously suppress a set $\mathcal{S}$ of skills we subtract the summed
vector:
\begin{equation}
\thedit = \thgoal - \alpha \cdot \sum_{i \in \mathcal{S}} \tauv_i,
\label{eq:multiskill}
\end{equation}
where, to hold the summed perturbation norm roughly fixed, the per-set scale
follows $\alpha \approx \alpha^\star_\text{single}/\sqrt{|\mathcal{S}|}$.

%%=============================================================
\section{Experiments}
\label{sec:experiments}

\subsection{Setup}
MergeVLA-LIBERO-Goal~\cite{fu2025mergevla}: a Qwen2.5-0.5B language backbone + dinosiglip-224px vision encoder + a $0.16$B action expert ($\approx\!1.4$B total; the ``0.68B'' tag counts the language backbone plus action expert, excluding the frozen vision encoder; see Supp.\ Implementation), per-task experts trained via LoRA rank~64. Benchmarks: \emph{LIBERO-Goal}~\cite{liu2023libero} (10 tasks), with \emph{-Spatial}, \emph{-Long10}, and \emph{-Object} for cross-suite tests; metric is closed-loop success rate (SR), 20 trials per evaluation unless noted; we report raw $x/n$ counts with 95\% \emph{Wilson} intervals~\cite{wilson1927} (which for $n{=}20$ are asymmetric and rate-dependent, e.g.\ a $0/20$ cell is $[0,16]\%$) rather than a single flat noise threshold, and use paired comparisons where the same initial states are reused across conditions. Baselines: Gradient Ascent (NegGrad) with $N \in \{500, 1000, 2000\}$ steps, and a norm-matched random-direction control confirming results are direction-specific.

\subsection{Initial Four-Skill Audit}
The Phase-1 negation matrix (Supp.\ Table~\ref{tab:phase1}) suppresses three of four targets to 0\%~SR at $\alpha=1.0$, with task~1 anomalously retained at 100\% (explained in Supp.\ Sec.~G). The random-direction control confirms suppression is direction-specific: the same magnitude in a random direction achieves only 25\% target reduction. Which component carries the edit? In the $t_0$ component ablation, VLM-only negation matches full negation whereas action-head-only negation leaves target-task SR at $80\%$, so \textbf{the VLM component is the primary driver of target suppression in this ablation}. Within the VLM, no single depth band suffices at its native norm: restricting the edit to any one band of the LM stack leaves the target largely intact (mean $91\%$ retained vs.\ $0\%$ at full depth; Supp.\ Table~\ref{tab:layerband}). Because each band carries a smaller-norm perturbation than the full vector, we read this as ``no tested single band suffices,'' not proof that skill identity is distributed across depth, so locate-then-edit methods~\cite{meng2023memit} targeting a few layers do not straightforwardly apply.

\subsection{Complete LIBERO-Goal Matrix: All Ten Skills}
\label{sec:extended}
\begin{figure*}[t]
\centering
\includegraphics[width=\textwidth]{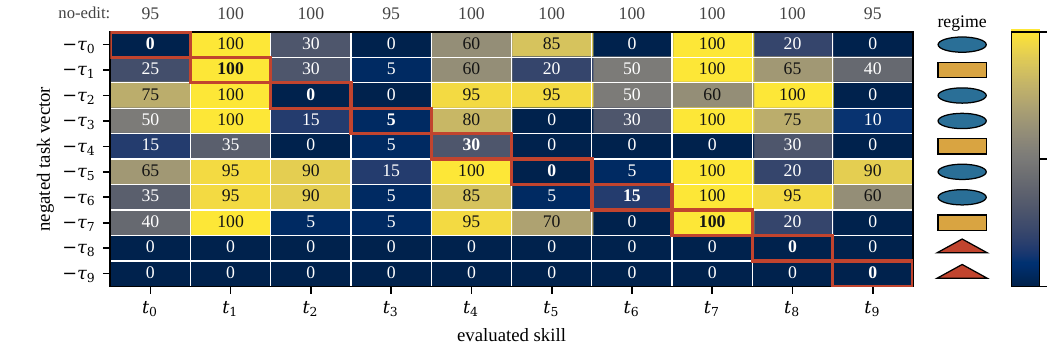}
\caption{\textbf{Full ten-skill audit reveals separation, resistance, and global
collapse.} Complete LIBERO-Goal negation matrix in natural task order (row $=$
negated $\tauv_i$, column $=$ evaluated skill; closed-loop SR\%, $20$ rollouts/cell,
$\alpha{=}1.0$; no-edit baselines above). Terracotta outlines mark the diagonal
\emph{target} cells (a low value indicates suppression; the box does not imply
success). The right strip labels each row's regime by color \emph{and} shape (ellipse $=$
\emph{separation}, rectangle $=$ \emph{resistant}, triangle $=$ \emph{global
collapse}): five \emph{target--control separation} rows (target suppressed, controls above the $40\%$ threshold),
three \emph{resistant} ($t_1,t_4,t_7$: target not suppressed), and two \emph{global
collapse} ($t_8,t_9$: target and controls both $\to\!0\%$). Target--control separation is
thus the outcome for half the Goal skills, not a universal property; supplement
tables give the normalized-retention statistics.}
\label{fig:goal10main}
\end{figure*}

We report the \emph{complete} $10\!\times\!10$ negation matrix over all ten Goal skills (Fig.~\ref{fig:goal10main}), each cell freshly rolled at 20 trials/cell in fixed task order. We use a single taxonomy throughout (also used for the cross-architecture tally, Supp.\ Table~\ref{tab:summary}, and the geometry scatter, Fig.~\ref{fig:collateral}C): a target is \emph{empirically suppressed} when its own-negation diagonal SR falls to $\leq\!20\%$ (a threshold fixed before analysis; suppressed skills here reach $\leq\!15\%$, so the split is insensitive to the cutoff); a row shows \emph{target--control separation} iff it is suppressed \emph{and} its mean off-diagonal control SR is $\geq\!40\%$ (raw); a row is \emph{resistant} if the target is not suppressed; and \emph{global collapse} if the target is suppressed but the mean control $<\!40\%$. The ten skills fall into three classes (Table~\ref{tab:regimes}): \textbf{five separation rows} ($t_0,t_2,t_3,t_5,t_6$: target $\leq\!15\%$, controls above the $40\%$ threshold, baseline-normalized retention $\GoalSepCRnorm$, range $\GoalSepCRrange$); \textbf{three resistant} ($t_1,t_7$ at $100\%$; $t_4$ borderline at $30\%$); and \textbf{two global collapses} ($t_8,t_9$: target $0\%$ but \emph{every} control also collapses), which we do not count as separation. This $5/3/2$ split is stable across separation thresholds $30$--$40\%$ (Supp.\ Table~\ref{tab:threshsens}) and its \emph{sign} reproduces across three seeds (Supp.\ Table~\ref{tab:multiseed}). Target--control separation is thus the outcome for half the skills, not a universal property even within this ``clean'' suite (a baseline-normalized selectivity index $\mathrm{Sel}_i$ quantifies the split, Supp.\ Table~\ref{tab:seli}).

On the five suppressible skills (each skill's negation vs.\ the mean of its own nine same-row controls, $n\!=\!5$ paired rows), negation drives \emph{targets} to a mean SR of $4.0\%$ while per-row \emph{controls} retain $57.3\%$, a $53.3$\,pp gap (row-bootstrap 95\% CI $[45.7,61.0]$, $d_z\!=\!5.3$, Cliff's $\delta\!=\!1.0$) with perfect row-level separation (every target $\leq\!15\%$ below its own control-row mean $\geq\!43.9\%$). Since the five-skill subset is selected post hoc (diagonal $\leq\!15\%$), we report this as a \emph{descriptive subgroup} effect, not primary confirmatory evidence; the post-hoc paired inferential tests are deferred to the supplement (Supp.\ \S\,B). The primary evidence is the full per-cell matrix (all ten Goal tasks) and the \piZeroFive{} dose--response. The split is not an $\alpha{=}1$ artifact but a dose--response (Supp.\ Table~\ref{tab:alpha_robust}): the five suppressible skills reach $\leq\!20\%$ by $\alpha{=}1.0$ while the three resistant skills resist until $\alpha\!\geq\!1.25$; by $\alpha{=}1.5$ the perturbation drives \emph{every} task to $0\%$, i.e.\ non-selective norm damage rather than a targeted edit.

\subsection{Constrained $\alpha$ Selection}
\label{sec:pareto}
Supp.\ Fig.~\ref{fig:pareto} shows target/control frontiers from sweeping $\alpha$. Under a fixed selection rule (smallest $\alpha$ with target SR $=0\%$, taking the one with highest mean control SR) the \emph{selected constrained operating point} for task~0 is $\alpha^\star = 0.75$ (target SR $=0\%$, mean control SR $\approx\!75\%$ on the three-control development panel $t_1$--$t_3$) versus $38\%$ on the same panel at the default $\alpha=1.0$, a large retention gain at identical target suppression. The same rule gives $\alpha^\star = 1.0$ for $t_2,t_3,t_5$ and $1.25$ for $t_6$ (Supp.\ Tables~\ref{tab:appendix_sweep},~\ref{tab:alpha_robust}; alpha ledger in the supplement); these are the values used in the held-out evaluation.

\paragraph{Held-out validation of the selected $\alpha^\star$ (full matrix).} Since $\alpha^\star$ is selected on rollouts, we re-evaluate each of the five suppressible skills at its selected $\alpha^\star$ on a \emph{disjoint} held-out set of initial states (indices $20$--$39$; selection used $0$--$19$), scoring the target \emph{and all nine controls} (Table~\ref{tab:heldoutmain}; full nine-control detail in Supp.\ Table~\ref{tab:heldout}), where the no-edit goal baselines remain high ($85$--$100\%$). Target suppression \emph{generalizes}: all five targets stay at $0\%$ ($0/20$). Selective control preservation is heterogeneous: mean baseline-normalized control retention ranges from $t_0$ ($78\%$) down to a near-total collapse for $t_6$ ($4\%$; mean $\HeldoutCRnorm$ across the five rows), and \emph{four of five} rows drive at least one control to $0\%$ (the fifth, $t_0$, still depresses its worst control to $35\%$ from a $\geq\!85\%$ baseline, so every row materially harms a nominally unrelated control). We therefore read single-vector negation as \emph{robust target suppression with fragile, heterogeneous locality}: the target-side effect transfers cleanly, the control-side selectivity only partly, so selection-panel margins are optimistic.

\begin{table}[t]
\centering
\caption{\textbf{Held-out validation} of each suppressible target at its selected
$\alpha^\star$, on initial states $20$--$39$ (disjoint from the $0$--$19$ used for
selection), $n{=}20$/cell. Every target is fully suppressed ($0/20$) yet
control locality is heterogeneous: mean baseline-normalized retention over the
nine controls falls from $78\%$ to $4\%$, and four of five rows zero at least one
control (worst-control). The $4\%$ floor is $t_6$ ($\alpha^\star{=}1.25$, control collapses). Reconciled with the Intro's $\GoalSepCRnorm$ (fixed $\alpha$): Supp.\ Table~\ref{tab:crosswalk}; per-control Table~\ref{tab:heldout}.}
\label{tab:heldoutmain}
\small
\setlength{\tabcolsep}{5pt}
\begin{tabular*}{\columnwidth}{@{\extracolsep{\fill}}lcccc@{}}
\toprule
Target & $\alpha^\star$ & Target succ. & Norm.\ ctrl.\ ret. & Worst ctrl. \\
\midrule
$t_0$ & $0.75$ & $0/20$ & $78\%$ & $35\%$ \\
$t_2$ & $1.0$  & $0/20$ & $65\%$ & $0\%$  \\
$t_3$ & $1.0$  & $0/20$ & $48\%$ & $0\%$  \\
$t_5$ & $1.0$  & $0/20$ & $65\%$ & $0\%$  \\
$t_6$ & $1.25$ & $0/20$ & $4\%$  & $0\%$  \\
\midrule
mean  &        & $0\%$  & $\HeldoutCRnorm$ & --- \\
\bottomrule
\end{tabular*}
\end{table}

\subsection{Failure of Selective Multi-Skill Subtraction}
Selective suppression does not survive composition. Cumulatively suppressing the five suppressible Goal skills at norm-matched $\alpha{=}1/\sqrt{k}$ and evaluating all eight (Supp.\ Table~\ref{tab:collapse_k}), the control skills fall to $10\%$ at $k{=}2$ and, for $k\!\geq\!3$, the policy collapses to $0\%$ on targets \emph{and} controls alike, even though these merge-typical scalings ($0.71$ down to $0.45$) sit below a single negation's $\alpha{=}1$. \emph{Sequential} suppression does not escape it: negating $t_0$ then $t_2$ at $\alpha{=}1$ collapses all four skills to $0\%$, including the control $t_1$ that single-$t_0$ negation had left at $100\%$. Nor do smarter rules: jointly negating $\{t_1,t_3\}$ at $\alpha{=}1.0$ drives both targets to $0\%$ but also collapses both controls $t_0,t_2$ (Supp.\ Table~\ref{tab:multiskill13}), and retain-orthogonal and mutually-orthogonal projected multi-negations of $\{t_0,t_2\}$ collapse the policy at $\alpha{=}1$ and retain the control at $\alpha{=}0.5$ only by leaving targets partially intact ($15$--$40\%$; Supp.\ Table~\ref{tab:projected}). Across the tested summation and projection rules, we observe reliable selectivity only in the single-direction, single-skill regime; we read this as an empirical boundary of the tested rules, not a general impossibility result.

\subsection{Single-Vector Sign vs.\ Composition}
\label{sec:direction}
\begin{figure}[t]
\centering
\includegraphics[width=\columnwidth]{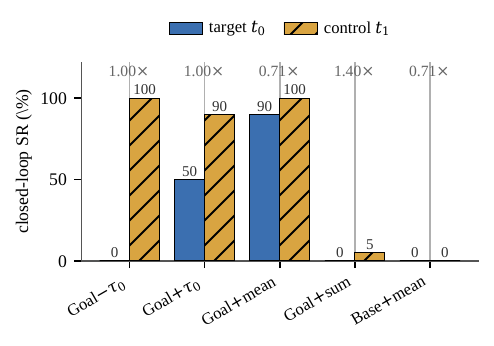}
\caption{\textbf{The tested $t_0$ outcomes depend on sign, anchor, and aggregate
scale.} Closed-loop SR of target $t_0$ and control $t_1$ (20 rollouts/cell; relative
perturbation norm $\times\lVert\tauv_0\rVert$ above each pair). Goal-anchored
subtraction shows separation ($0/100$); the equal-norm addition is retained ($50/90$);
the $0.71\times$ goal-anchored mean is retained ($90/100$). The $1.4\times$
goal-anchored sum nearly collapses the pair ($0/5$), whereas the base-anchored mean
collapses both skills ($0/0$); full per-condition values are in Supp.\ Table~\ref{tab:direction}.}
\label{fig:direction_main}
\end{figure}

Anchored at $\thgoal$ and held at matched norm ($1.0\times$, $\lVert\tauv_0\rVert\!\approx\!83$), \emph{subtracting} $\tauv_0$ drives its own task to $0\%$ (control $100\%$), whereas \emph{adding} the identical vector leaves the task at $50\%$ (control $90\%$; Supp.\ Table~\ref{tab:direction}). Composition depends on anchor and aggregate scale: onto $\thgoal$, a mean $\tfrac12(\tauv_0{+}\tauv_1)$ at $0.71\times$ is retained ($90\%/100\%$) while the $1.4\times$ sum collapses ($0\%/5\%$). The same $0.71\times$ mean anchored at the base $\thbase$, from which a single vector still injects its skill ($95\%$), also collapses ($0\%/0\%$). At matched norm around this Goal anchor the probe thus shows a direction-specific sign asymmetry (Fig.~\ref{fig:direction_main}): the equal-norm \emph{addition} leaves the target at $50\%$ and a norm-matched \emph{random-direction} edit achieves only ${\sim}25\%$ target reduction (\S\,Setup), so the suppression is not generic norm damage but specific to the negated expert direction. Sparsifier and LoRA-rank sweeps are consistent (Supp.\ Table~\ref{tab:rank}).

\begin{figure*}[t!]
\centering
\includegraphics[width=\textwidth]{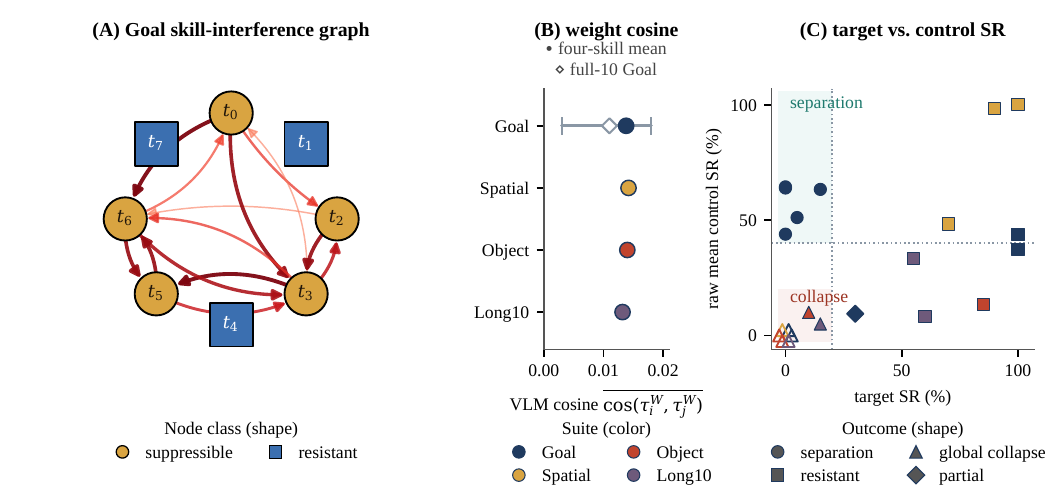}
\caption{\textbf{Collateral damage is structured, whereas mean weight-space
cosine does not track control survival.} \emph{(A)} Directed Goal interference graph on the
$8$-skill core (excluding the two global-collapse rows $t_8,t_9$): each edge
$i\!\to\!j$ marks a $\geq\!50$-point control-SR drop from subtracting $\tauv_i$,
drawn only for source rows whose own target is suppressed. Resistant skills thus
have \emph{no} outgoing edges by construction: not because negating them is
harmless, but because it fails to suppress their target and damages controls only
via non-selective collapse (Supp.\ Table~\ref{tab:seli}). Edge width/shade scale
with the drop; node shape/color mark suppressible vs.\ resistant ($t_3$ is the
largest collateral hub); the full $10\!\times\!10$ matrix is Fig.~\ref{fig:goal10main}.
\emph{(B)} Mean VLM task-vector cosine per suite: filled dots are the matched
four-skill probe ($6$ pairs each), and the Goal open diamond/whisker is the full
ten-skill distribution ($90$ ordered pairs; mean $0.011$). \emph{(C)} Target SR
versus raw mean control SR for every negated row across the four LIBERO suites;
color denotes suite, shape denotes taxonomy outcome, and dotted lines mark the
$20\%/40\%$ thresholds. Goal alone densely occupies the separation quadrant.}
\label{fig:collateral}
\end{figure*}

\subsection{Comparison with Gradient Ascent}
\label{sec:ga}
We compare negation against three data-dependent gradient baselines and a leave-one-task-out retraining oracle on a \emph{matched panel} (same target $t_0$, same nine controls, same held-out states, $n{=}20$/cell; Table~\ref{tab:e1}). All drive the target to $0/20$. Plain NegGrad~\cite{graves2021amnesiac} collapses \emph{all nine controls} to $0\%$; \textbf{Gradient-Difference}~\cite{liu2022continual} retains $88.3\%$ raw mean control SR, exceeding negation ($73.9\%$); \textbf{NegGrad+}~\cite{kurmanji2023scrub} retains $66.1\%$ (Wilson $[59,73]\%$) and zeroes one control. The retraining oracle is the upper bound ($95.6\%$ mean control) but requires a full retrain, whereas negation provides $73.9\%$ raw matched-panel control SR with no removal-time data or gradient steps.

\begin{table}[t]
\centering
\caption{\textbf{Matched-panel comparison} (target $t_0$; identical nine-control
panel $t_1$--$t_9$; identical held-out initial states, offset $20$, disjoint from
$\alpha^\star$ selection; $n{=}20$/cell). All five removal methods suppress the target ($0/20$).
Retain-aware optimization does \emph{not} uniformly beat the data-free edit:
Gradient-Difference exceeds negation, NegGrad+ does not (and zeroes one control).
Per-control counts, Wilson CIs, and paired outcomes: Supp.\ Table~\ref{tab:ga}.}
\label{tab:e1}
\small
\setlength{\tabcolsep}{3pt}
\begin{tabular*}{\columnwidth}{@{}l c c c l@{}}
\toprule
Method & Target & Mean ctrl & Worst ctrl & Removal-time need \\
\midrule
No-edit           & $20/20$ & $93.3\%$ & $85\%$ & --- \\
\textbf{Negation} & $0/20$  & $73.9\%$ & $35\%$ & \textbf{none} \\
NegGrad           & $0/20$  & $0.0\%$  & $0\%$  & forget, grad \\
NegGrad+          & $0/20$  & $66.1\%$ & $0\%$  & forget/retain, grad \\
Grad-Diff         & $0/20$  & $88.3\%$ & $35\%$ & forget/retain, grad \\
Retrain oracle    & $0/20$  & $95.6\%$ & $85\%$ & full retrain \\
\bottomrule
\end{tabular*}
\end{table}

\subsection{Generality and Boundaries of the Effect}
\label{sec:objlong}

\paragraph{Collateral Structure Is Not Explained by Mean Cosine.}
Collateral damage is structured (Fig.~\ref{fig:collateral}A): the five suppressible Goal skills form a dense directed interference cluster, with $t_3$ the largest hub. Yet matched four-skill VLM cosine means are $0.0132$--$0.0142$ across suites (full ten-skill Goal mean $0.011$ over $90$ ordered pairs), while off-diagonal SR spans $5.8$--$61.7\%$ (Fig.~\ref{fig:collateral}B--C). Neither tested forward proxy explains cross-suite control survival. Behavioral overlap remains a hypothesis, not an established mechanism. Consistent with this limited scope, the alignment score ranks suppressibility \emph{within} a suite, but its threshold does not transfer \emph{across} suites (Supp.\ Sec.~E). Per-suite facets and the full architecture$\times$suite tally are in Supp.\ Fig.~\ref{fig:geom_selectivity} and Supp.\ Table~\ref{tab:summary}.

\paragraph{Cross-suite: where the effect breaks.}
\label{sec:spatial}
On Spatial, no row shows separation: only $t_0$ is fully suppressed and its controls collapse; $t_2$ is partial, whereas $t_1$ and $t_3$ resist (Supp.\ Table~\ref{tab:spatial}). Object suppresses $3/4$ targets but retains only $5.8\%$ off-diagonal SR; Long10 suppresses $2/4$ with $11.7\%$ off-diagonal SR (Supp.\ Tables~\ref{tab:object}--\ref{tab:long10}). Selectivity is therefore strongest on Goal and fails on the tested suites with shared behaviors, despite similar mean weight-space cosine (Supp.\ Table~\ref{tab:summary}); Spatial's $61.7\%$ off-diagonal SR is inflated by resistant rows rather than genuine locality.

%%=============================================================
% Table 3 declared before Fig 5 (given [t]) so it takes the cross-architecture
% page top while the full-width Fig 5 spans the following column region; the
% expanded OpenVLA paragraph then fills the left column with no mid-column gap.
% This keeps the Discussion section whole (Gate 1B) without adding vspace/pages.
\begin{table}[t]
\centering
\caption{\textbf{Three closed-loop regimes on LIBERO-Goal} (decision rule: own-negation diagonal SR; mean off-diagonal control
SR). Only \emph{target--control separation} yields suppression with retained controls.}
\label{tab:regimes}
\small
\setlength{\tabcolsep}{4pt}
\begin{tabular*}{\columnwidth}{@{\extracolsep{\fill}}lccl@{}}
\toprule
Regime & Target SR & Mean ctrl SR & Goal skills \\
\midrule
Separation      & $\leq\!20\%$ & $\geq\!40\%$ & $t_0,t_2,t_3,t_5,t_6$ \\
Resistant       & $>\!20\%$    & ---          & $t_1,t_4,t_7$ \\
Global collapse & $\leq\!20\%$ & $<\!40\%$    & $t_8,t_9$ \\
\bottomrule
\end{tabular*}
\end{table}

\begin{figure*}[t!]
\centering
\includegraphics[width=0.92\textwidth]{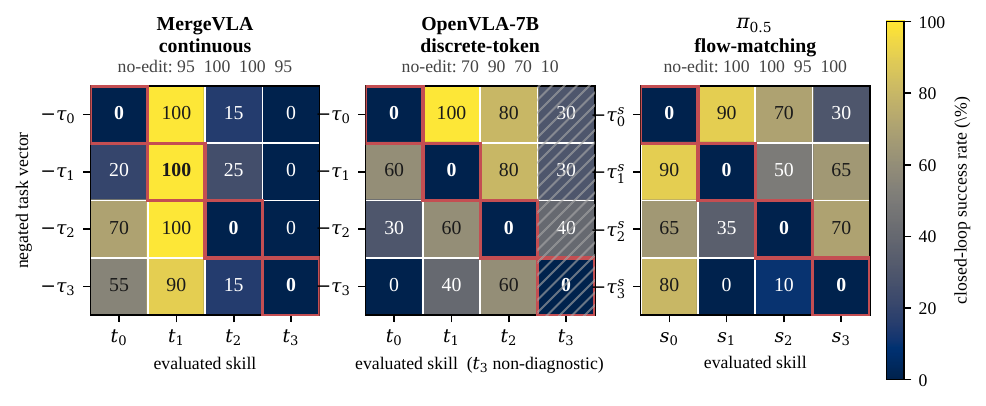}
\caption{\textbf{Target suppression recurs across three action-head families, but
control retention remains model-dependent.} Rows subtract $\tauv_i$; columns
evaluate skills; red boxes mark target cells, and no-edit baselines appear above
each matrix. MergeVLA $t_1$ resists suppression; OpenVLA $t_3$ is hatched because
its baseline is $10\%$. Rollouts/cell: MergeVLA $20$, OpenVLA $10$ (diagonals
$20$), and \piZeroFive{} $20$. The \piZeroFive{} panel uses four different Goal
tasks (mapping in Supp.\ Table~\ref{tab:taskmap}). Absolute off-diagonal margins
remain model-dependent at these rollout counts.}
\label{fig:dualheat}
\end{figure*}

\paragraph{Cross-architecture: OpenVLA-7B.}
\label{sec:openvla}
On \textbf{OpenVLA-7B}~\cite{kim2024openvla} (a discrete-token, Open-X-pretrained, $\sim$10$\times$ larger backbone), negation suppresses all three high-baseline diagnostic targets ($t_0,t_1,t_2$) to $0\%$, while $t_3$ is non-diagnostic because its no-edit success rate is only $10\%$ and cannot demonstrate suppression. The retained controls, however, are far less protected than on the continuous Goal model: mean off-diagonal SR is $50.8\%$ across all twelve off-diagonal cells ($10$ rollouts/cell, diagonals rerun at $20$; Supp.\ Table~\ref{tab:openvla}; Fig.~\ref{fig:dualheat}), and adding all four task vectors at once collapses every skill, mirroring the multi-skill failure we observe on the continuous backbone. That the same qualitative target--control pattern recurs despite substantial differences in backbone scale, pretraining corpus, and action representation supports reading it as a transferable property of per-skill negation rather than an artifact of one architecture; the precise cell-wise margins, by contrast, remain model-dependent at these rollout counts.

%%=============================================================
% Gate 1B: flush Fig 5 (fig:dualheat) before Discussion so the section is not
% split across the float. AAAI-safe (placeins, no hyperref); no vspace/margin change.
\FloatBarrier
\section{Discussion and Limitations}
\label{sec:limitations}

\begin{figure*}[t]
\centering
\includegraphics[width=\textwidth]{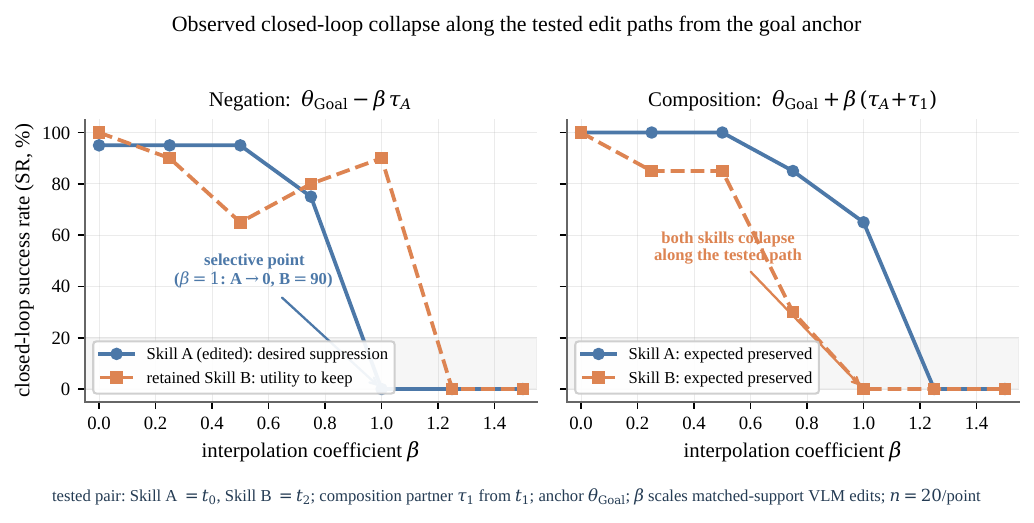}
\caption{\textbf{Observed closed-loop collapse along the tested edit paths from
the goal anchor.} Closed-loop success rate of the edited Skill~A ($t_0$, blue
circles) and the retained Skill~B ($t_2$, orange squares) as we interpolate from
$\thgoal$ along the negation direction $-\tauv_A$ (\emph{left}) and the
composition direction $+(\tauv_A{+}\tauv_1)$ (partner $\tauv$ from $t_1$)
(\emph{right}); $20$ rollouts/point, single tested pair. \emph{Left} (desired:
suppress~A, keep~B): negation reaches a \textbf{selective operating point} at
$\beta{=}1.0$: A$\to\!0\%$, B$=90\%$. \emph{Right} (expected: preserve both
skills): both collapse, with B falling \emph{before} A, so the target is only
suppressed after the other skill is already suppressed. The five-pair population
is in Supp.\ Table~\ref{tab:barrier_pairs}, and the accompanying static-loss and
sharpness measurements are correlational (Supp.\ \S\,G).}
\label{fig:barrier}
\end{figure*}

\paragraph{Diagnostics and what we do \emph{not} claim.}
Exploratory sharpness, path, alignment, and localization probes co-vary with the outcomes but do not establish a mechanism (Supp.\ \S\S\,D--H). What they do show is that the negation and composition edits behave differently in closed loop along the paths we test from the Goal anchor (Fig.~\ref{fig:barrier}): negation descends into a selective plateau where the target falls to $0\%$ while the control skill is retained at $90\%$, whereas composition never reaches such a point and collapses the control skill \emph{before} the target. Crucially this asymmetry is behavioral rather than a static-loss artifact, since the composition path that collapses the policy raises the teacher-forced loss \emph{less} than the negation path that does not. We read these diagnostics as characterizing, not explaining, the effect: negation is temporary behavioral suppression, not certified or safety-grade unlearning~\cite{li2024wmdp}, and a practitioner should not treat a closed-form subtraction as a permanence guarantee.

\paragraph{Limitations.} Five scope limits bound our claims.
(1)~\emph{Single training seed}: the full $10\!\times\!10$ matrices use one training seed; a three-seed $t_0$/$t_1$ probe reproduces the target--control gap with a run-dependent threshold (Supp.\ Table~\ref{tab:multiseed}), but we do not claim seed robustness for the matrices.
(2)~\emph{Simulation only}: all evaluation is LIBERO MuJoCo with Wilson intervals (per-panel rollout counts in Supp.\ \S\,A); real-robot and non-LIBERO validation remain gaps.
(3)~\emph{Expert-delta dependency}: clean single-skill selectivity presumes per-skill expert banks we train; negating an off-the-shelf multitask vector is broadly destructive.
(4)~\emph{Cross-suite failure}: selectivity breaks on Object and Long-horizon (controls collapse with the target), so the sharpest claims are calibrated on MergeVLA-Goal.
(5)~\emph{Baseline-comparison scope}: the matched E1 panel covers one target ($t_0$), one seed, and one budget per gradient method; with heterogeneous controls on fixed states, pooled intervals are descriptive and we read E1 as a matched cost--retention comparison, not an architecture-wide ranking.

\paragraph{Relearning: consistent with masking.}
If subtraction genuinely erased the skill from the weights, re-acquiring it should be no easier than learning it from a model that never had it. We therefore fine-tune, under a pre-specified rule, the negated $t_0$ checkpoint against a task-naive base that never saw $t_0$, on the same fixed initial states (Supp.\ Table~\ref{tab:relearn}). The recovery slopes diverge sharply: by $k{=}250$ steps the negated model is already back to $85\%$ ($17/20$, Wilson $[64,95]\%$) while the task-naive floor is still at $0\%$ ($0/20$, Wilson $[0,16]\%$), non-overlapping intervals. The suppressed skill is thus recoverable far faster than it is learnable from scratch, which is what a quickly reversible behavioral mask predicts rather than durable weight-level deletion. This single-skill, single-seed probe does not rule out all forms of forgetting, so we do not claim a permanence result and instead treat rapid recovery under brief re-exposure as a genuine failure mode that a target-only metric would silently miss.

\paragraph{Why closed loop matters.}
In an embodied policy an edit changes not only the current action but the future observations it induces, so a perturbation that looks local under one-step loss can accumulate into a control failure over a rollout. Locality must therefore be evaluated behaviorally, on a held-out target-by-control panel, against pre-specified thresholds on target SR, mean normalized control retention, and worst-control retention.

\paragraph{Minimum reporting unit.}
For each edit we recommend reporting target $x/n$, every control $x/n$, baseline-normalized mean and worst-control retention, the selection/evaluation split, and the removal-time data and optimization required. Resistant rows should not enter a locality average, since their high control scores coexist with a failure to suppress the target. These are audit quantities, not safety guarantees.

% Conclusion-adjacent single-column figure: turns the closing "78% down to 4%"
% sentence into the visible retention cliff. Numbers are the held-out
% baseline-normalized nine-control retention and worst-control from
% generated/results_macros.tex (\HeldoutCRnormEach / \HeldoutWorstEach), i.e.
% the same values as Table~\ref{tab:heldoutmain}. Rendered INLINE (non-floating,
% via \captionof) between the Minimum-reporting paragraph and the Conclusion so
% the figure sits ABOVE the Conclusion text in reading order, per the intended
% layout; a floating [b] would instead push it to the column bottom below the
% Conclusion. No manual vspace/margin.
\par\medskip
\begingroup
\centering
\includegraphics[width=\columnwidth]{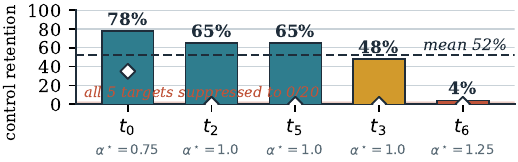}
\captionof{figure}{\textbf{Locality is a cliff, not a floor.} Held-out control
retention across the five suppressible Goal targets at their selected
$\alpha^\star$ (bars: baseline-normalized nine-control mean; diamonds: worst
single control); values match Table~\ref{tab:heldoutmain}.}
\label{fig:heldoutcliff}
\par
\endgroup
\medskip

\section{Conclusion}

We audit task-vector subtraction through closed-loop target and control behavior. It rapidly suppresses some VLA skills, but control retention is heterogeneous: the pattern recurs across continuous, discrete-token, and flow-matching policies yet breaks on entangled suites and multi-skill edits, and a matched-norm probe shows a local sign asymmetry around one Goal anchor. A target-only metric would score all five held-out edits successful, yet the full control panel reveals retention spanning $78\%$ down to $4\%$ (Table~\ref{tab:heldoutmain}; visualized in Fig.~\ref{fig:heldoutcliff}), so closed-loop target and control evaluation is essential for locality in embodied model editing.

%%=============================================================
% Start References on a fresh page so body+figures occupy pp.1-7 and
% References are confined to pp.8-9 (page-break only; no font/margin/spacing change).
% \raggedbottom on the reference pages so the shorter left column is NOT vertically
% stretched to match the right (removes the large inter-bibitem gaps); no \bibsep,
% font, or margin change.
\clearpage
\raggedbottom
% Small emergencystretch helps the bibliography line-breaker balance the two
% reference columns without changing font, margins, or inter-bibitem spacing.
\setlength{\emergencystretch}{2em}
\bibliography{aaai2027}

\ifmain\else
\clearpage
\appendix
% Compact supplement header for the arXiv single-file build (the AAAI split build
% carries an equivalent header in supp.tex). Not a separate cover page: the
% appendix text flows immediately below it.
\begin{center}
{\large\bfseries Technical Supplement}\\[2pt]
{\itshape Suppression Sticks, Locality Doesn't:\\
A Closed-Loop Target-and-Control Audit of Task-Vector Negation
in VLA Policies}
\end{center}
\vspace{4pt}
% S-number appendix floats so body '(Supp. Table Sx)' pointers match
\renewcommand{\thetable}{S\arabic{table}}
\renewcommand{\thefigure}{S\arabic{figure}}
\renewcommand{\thelisting}{S\arabic{listing}}
\setcounter{table}{0}\setcounter{figure}{0}\setcounter{listing}{0}

\section{A. Implementation Details and Metric Definitions}

This appendix restates the negation operator and the Alignment Score $\mathrm{AS}_i$ in full, then documents every implementation and evaluation detail needed to reproduce the results. Unless noted otherwise, all success rates are over 20 rollout trials per (task, condition) cell using the protocol below.

\paragraph{Task identifiers.} Table~\ref{tab:taskmap} maps every task index used in the paper to its natural-language instruction. LIBERO-Goal skills are $t_0$--$t_9$; the flow-matching \piZeroFive{} panels use $s_0$--$s_3$, which are four of the same LIBERO-Goal tasks (their $t$-index given in the table).

\begin{table}[htbp]
\centering
\caption{\textbf{Task-index to instruction mapping.} LIBERO-Goal ($t_0$--$t_9$)
and the \piZeroFive{} Goal panel labels $s_0$--$s_3$ (with their LIBERO-Goal
$t$-index).}
\label{tab:taskmap}
\small
\setlength{\tabcolsep}{5pt}
\begin{tabular}{ll}
\toprule
ID & Instruction (\texttt{libero\_goal}) \\
\midrule
$t_0$ & open the middle drawer of the cabinet \\
$t_1$ & put the bowl on the stove \\
$t_2$ & put the wine bottle on top of the cabinet \\
$t_3$ & open the top drawer and put the bowl inside \\
$t_4$ & put the bowl on top of the cabinet \\
$t_5$ & push the plate to the front of the stove \\
$t_6$ & put the cream cheese in the bowl \\
$t_7$ & turn on the stove \\
$t_8$ & put the bowl on the plate \\
$t_9$ & put the wine bottle on the rack \\
\midrule
$s_0$ ($t_7$) & turn on the stove \\
$s_1$ ($t_8$) & put the bowl on the plate \\
$s_2$ ($t_1$) & put the bowl on the stove \\
$s_3$ ($t_2$) & put the wine bottle on top of the cabinet \\
\bottomrule
\end{tabular}
\end{table}

\paragraph{The method in full.} Listing~\ref{lst:core} gives the entire method (the closed-form negation edit and the forward-only Alignment Score $\mathrm{AS}_i$) in a few lines.

\begin{listing}[htbp]
\begin{lstlisting}[language=Python]
# Target suppression: one closed-form weight edit (< 1 s, CPU).
tau_i   = expert_i - base            # skill vector (per-tensor)
theta_e = theta_goal - alpha * tau_i # subtract update: suppress target behavior i

# Forward-only Alignment Score AS_i (no negation outcome used):
def AS(i, eps=0.3):
    h0 = hidden(theta_goal, D[i])              # baseline reps on skill-i data
    hi = hidden(theta_goal - eps*tau_i, D[i])  # own-skill shift
    own = norm(hi - h0)
    others = mean([norm(hidden(theta_goal - eps*tau_i, D[k])
                        - hidden(theta_goal, D[k])) for k in tasks if k != i])
    return own / others                        # > 1  => target i predicted suppressible
\end{lstlisting}
\caption{The complete method: negation is a single weight subtraction, and
the a~priori Alignment Score $\mathrm{AS}_i$
is a handful of forward passes on the goal model. The edit itself is data-free
and gradient-free at application time (given precomputed task vectors); the
Alignment Score is gradient-free and forward-only but \emph{data-dependent}
(it reads skill-$i$ inputs). Scripts and tabular data used to reproduce the
reported analyses are included in the Code and Data Supplement.}
\label{lst:core}
\end{listing}

\paragraph{Implementation details.} \textbf{Models and parameter accounting.} The goal model is the MergeVLA-LIBERO multitask
checkpoint (Prismatic-style VLA: Qwen2.5-0.5B language backbone +
dinosiglip-224px vision encoder + a $0.16$B action expert and a two-layer
proprio projector). Its parameter budget, measured from the checkpoint tensors, is:
language model $0.49$B, vision encoder $0.73$B, projector $0.03$B, action expert
$0.16$B; so the \emph{full} VLM weight space (language$+$vision$+$projector) is
$1.25\times10^{9}$ params and the whole policy is $\approx\!1.41$B. The tag
``$0.68$B'' used as the model nickname counts the \emph{language backbone plus
action expert} ($0.49{+}0.16{+}0.03\!\approx\!0.68$B), i.e.\ everything except the
frozen dinosiglip vision encoder; we retain it only as a model identifier. The
\emph{edited support} is smaller still: per-task experts are LoRA fine-tunes over
the language-model and action-head weights only (rank 64, $\alpha_{\text{LoRA}}=128$,
dropout 0; the $0.73$B vision encoder is frozen, so its reconstructed task-vector
entries are $\approx\!0$). The full-VLM cosine geometry (Fig.~\ref{fig:collateral}B,
Fig.~\ref{fig:geometry}) is nonetheless computed over the complete $1.25\times10^{9}$
VLM weight space. Experts are trained from a shared initialization $\thbase$ for 10{,}000
steps with AdamW, learning rate $2\!\times\!10^{-4}$, batch size 8 with
2-step gradient accumulation, and \texttt{num\_images\_in\_input}$=2$.
The action head and proprio projector use a fixed shared initialization
(seed 42) across all experts so that cross-task differences are a function
of data alone.

\textbf{Skill vectors.} $\tauv_i = \theta_{\mathrm{exp},i} - \thbase$ is
formed component-wise over VLM weights, action head, and proprio
projector. Negation $\thedit = \thgoal - \alpha\tauv_i$ is a single CPU
tensor operation ($<1$\,s); no data or gradient is used.

\textbf{\piZeroFive{} per-skill experts.} For the flow-matching
per-skill matrices (Table~\ref{tab:pi05matrix}) we full-fine-tune (not LoRA)
the public \texttt{lerobot/pi05\_libero\_base} on each single skill's
demonstrations for $8$k steps (AdamW, batch size~1 with gradient
checkpointing, vision encoder frozen, VLM \emph{unfrozen} so the per-skill
$\tauv_i$ captures language-model adaptation); $\tauv_i$ is then negated on
the public multitask \texttt{pi05\_libero\_finetuned} policy. We report
$\lVert\tauv_i\rVert\!\approx\!55$--$57$ across skills, vs.\
$\lVert\tauv_{\mathrm{multitask}}\rVert\!=\!242$.

\textbf{Evaluation.} Closed-loop LIBERO rollouts with the MuJoCo/EGL
renderer, 20 trials per (model, task) cell unless noted, success defined
by the benchmark's task-completion predicate.

\textbf{Compute.} Training and evaluation run on NVIDIA A800-80GB GPUs; experts
train in ${\sim}6$--$18$\,h each depending on data-loader contention. Software:
PyTorch 2.2.0 built against the CUDA 12.1 toolkit, running on a host driver/runtime
that reports CUDA 12.8 (the newer driver runs the cu121 build); the
MergeVLA/Prismatic stack with FlashAttention-2; and
\texttt{tensorflow-datasets} for the RLDS data pipeline. All experts share
the seed-42 shared initialization (\S A) so cross-task differences are a
function of data, not initialization.

\paragraph{Metric definitions.} \textbf{Sampled perturbation sensitivity.} For
parameters $\theta$ and a fixed mini-batch, $\mathrm{sens}_\varepsilon(\theta) =
\max_n [L(\theta+\delta_n)-L(\theta)]/L(\theta)$ over $n{=}20$ Gaussian
perturbations $\delta_n$. This is a \emph{sampled} maximum, not a true worst-case
inner maximization, so we call it sampled perturbation sensitivity rather than
$\varepsilon$-sharpness; we retain the shorthand ``sharpness'' loosely below. The \emph{absolute} variant draws
$\delta\sim\mathcal{N}(0,\varepsilon^2 I)$; the \emph{adaptive (ASAM)}
variant scales each coordinate by $|\theta|$; the \emph{filter-normalized}
variant scales by per-tensor $\lVert\theta\rVert/\sqrt{\dim}$.

\textbf{Alignment Score $\mathrm{AS}_i$, a forward-only target-suppression susceptibility diagnostic.} For skill $i$,
$\mathrm{AS}_i = \mathbb{E}_{x\sim\mathcal{D}_i}\lVert h(\thgoal-\epsilon\tauv_i;x)-h(\thgoal;x)\rVert
\,/\,\frac{1}{N-1}\sum_{k\neq i}\mathbb{E}_{x\sim\mathcal{D}_k}\lVert h(\thgoal-\epsilon\tauv_i;x)-h(\thgoal;x)\rVert$,
where $h(\cdot;x)$ is the final VLM hidden state. It uses only the goal
model and $\tauv_i$ (no negation outcome); we report it averaged over
$\epsilon\in[0.1,0.5]$.

\section{B. Extended Negation Results and the $\alpha$ Trade-off}
\paragraph{Negation matrices and multi-skill collapse.} Table~\ref{tab:phase1} gives the Phase-1 $4\times4$ LIBERO-Goal matrix at $\alpha=1.0$, in which the bold diagonal marks the target skill each negated vector is meant to erase; the lone $\dagger$ on Task~1 flags a direction-misalignment anomaly we dissect in Supp.\ Sec.~G. Scaling to the complete ten-skill matrix (main paper), the skills split into five selective, three resistant, and two non-selective global collapses, showing the effect is neither an artifact of the small probe nor uniform across skills. Table~\ref{tab:effect} re-expresses these outcomes as relative SR drops $e_{ij}=1-\mathrm{SR}_{ij}/\mathrm{SR}^*_j$, isolating the diagonal target effects from off-diagonal spillover so that erasure and retention are read on a common normalized scale.

\paragraph{Post-hoc subgroup inferential tests (five suppressible rows).} For completeness we record the paired inferential statistics for the post-hoc five-suppressible-skill subgroup (diagonal $\leq\!15\%$) whose descriptive nine-control gap ($53.3$\,pp; row-bootstrap 95\% CI $[45.7,61.0]$; $d_z\!=\!5.3$; Cliff's $\delta\!=\!1.0$) is reported in the main text; all derived statistics here are recomputed from the canonical ten-skill matrix with the consistent nine-control denominator (\texttt{recompute\_ninecontrol\_stats.py}, seed $20260725$, $10^5$ row bootstraps). Because the subgroup is selected post hoc and the rows share fixed initial states, these are secondary, conditioned-on-selection statistics: paired $t$-test $p\!=\!2.9\!\times\!10^{-4}$ and one-sided paired Wilcoxon $p\!=\!0.031$ (both signed-rank $p$-values sit at their mechanical $n\!=\!5$ floors (one-sided $1/32\!=\!0.03125$, two-sided $2/32\!=\!0.0625$) because the post-hoc subgroup has all five control gaps positive (Cliff's $\delta\!=\!1.0$), so these are conservative bounds, not informative significance levels), corroborated by the \piZeroFive{} dose-response (Fisher $p\!=\!2.6\!\times\!10^{-8}$). We do not treat these $p$-values as primary confirmatory evidence; the primary evidence remains the full ten-task per-cell matrix together with the flow-matching dose-response.

\paragraph{The complete $10\!\times\!10$ matrix, with no reordering.} To avoid selective row presentation, the \emph{full} $10\!\times\!10$ LIBERO-Goal negation matrix is the main-paper Figure~\ref{fig:goal10main}: every cell freshly rolled at $20$ trials, tasks in fixed index order, diagonal boxed. The two new skills complete the picture: negating $t_8$ or $t_9$ (Goal baselines $100\%$ and $95\%$, so genuinely learned) drives the target to $0\%$ but \emph{also} collapses every control to $0\%$ (the two triangle-coded global-collapse rows in Fig.~\ref{fig:goal10main}). These are within-suite instances of the same non-selective failure we document across the entangled Object/Long10 suites: negation \emph{fires} but is not \emph{selective}. The honest aggregate is therefore the raw off-diagonal over all ten rows, $\GoalAllOffdiag$; the genuine locality lives only in the five separation rows, where baseline-normalized control retention is $\GoalSepCRnorm$ (range $\GoalSepCRrange$). We deliberately do \emph{not} report an ``eight-row'' mean: the three resistant rows keep their controls high only because their target is never suppressed, so folding them in would inflate an apparent locality that the edit does not have. Selective suppression is thus the rule for half of the Goal skills but not universal even within a ``clean'' suite, consistent with the behavioral-separability boundary as the governing factor rather than the suite label.

\paragraph{Multi-skill collapse.} Table~\ref{tab:collapse_k} pushes negation beyond a single skill, cumulatively suppressing $k$ of the five suppressible Goal skills at $\alpha=1/\sqrt{k}$ to hold the summed perturbation norm roughly fixed. Targets continue to be suppressed, but control retention collapses at $k\!\geq\!2$: selective suppression does not survive the superposition of multiple negation vectors. This is a boundary of the method: it is a single-skill edit with limited selectivity, not a composable multi-skill one.

\paragraph{Suppression is a skill-level edit, not an instruction-template trigger.} A natural worry is that negation merely detaches the policy from one \emph{phrasing} of the task rather than removing the underlying skill. We test this directly on the suppressed skill $t_0$ (``open the middle drawer of the cabinet''). The goal model executes three semantic-preserving paraphrases at $90/80/95\%$ (canonical $95\%$), confirming the paraphrases are intelligible and executable; the negated model fails \emph{all three} at $0\%$, exactly matching its $0\%$ on the canonical instruction. Because these three tested wordings do not recover the skill, the suppression was not bypassed by the tested paraphrases, consistent with the edit operating at the level of the behavior rather than one instruction template. We do not test adversarial or automatically searched rephrasings, so this is evidence against template-triggering for the tested phrasings, not a general claim that suppression cannot be bypassed by rephrasing.

\paragraph{The $\alpha$ trade-off.} Table~\ref{tab:appendix_sweep} sweeps $\alpha$ per task, reporting target SR ($\downarrow$) against mean control SR ($\uparrow$) and bolding the selected constrained operating point $\alpha^\star$ (max control SR s.t.\ target SR${=}0$). Table~\ref{tab:alpha_robust} shows the suppressible skills are suppressed across a broad $\alpha$ band, whereas resistant skills hold until $\alpha\!\geq\!1.25$ ($t_4$ already at $0\%$ there) and reach $0\%$ collectively by $\alpha=1.5$, where \emph{all} skills collapse non-selectively, so their apparent suppression is not selective. Table~\ref{tab:pareto} and Figure~\ref{fig:pareto} trace the frontier: at every selected operating point target SR $=0\%$ ($0/20$ observed; Wilson upper bound ${\sim}16\%$) while controls are retained, recovering substantial control retention over the fixed-$\alpha{=}1.0$ default (e.g.\ Task-0 $75\%$ vs.\ $38\%$ on the three-control development panel $t_1$--$t_3$); Task~1's diamond is annotated with its direction-misalignment anomaly in the figure.

\paragraph{Direction and magnitude.} Table~\ref{tab:direction} and Figure~\ref{fig:direction} norm-match edits by their $\ell_2$ weight-space perturbation relative to $\lVert\tauv_i\rVert$. For a \emph{single} vector at matched norm ($1.0\times$), sign is decisive: negation $-\tauv_0$ is selectively suppressive (target $0\%$, control $100\%$) while the equal-norm addition $+\tauv_0$ is retained (target $50\%$, control $90\%$): the asymmetry here is set by direction, not magnitude. For \emph{multi-skill} composition the outcome instead tracks the anchor and aggregate magnitude: onto $\thgoal$ the $0.71\times$ mean composition is retained ($90\%/100\%$) but the $1.4\times$ sum collapses ($0\%/5\%$), and the same $0.71\times$ mean anchored at $\thbase$ collapses ($0\%/0\%$) even though a single vector injects cleanly there ($95\%$). This base-anchored multi-skill collapse is not specific to naive averaging: TIES ($30\%$ density, sign-elected) and DARE ($90\%$ drop, survivor-rescaled) composition of $\{t_0,t_1\}$ from $\thbase$ both also give $0\%$ on both composed skills. On the tested $t_0$ probe, sign governs the single-vector edit; anchor and magnitude govern composition, across standard merge methods. Sparsifier and rank sweeps are consistent with the direction account: single-vector TIES-negation stays selective, DARE collapses the target only under $3.16\times$ over-scaling, and retention is robust across LoRA ranks $\{16,32,64\}$ (Table~\ref{tab:rank}).

\begin{table}[htbp]
\centering
\caption{\textbf{LIBERO-Goal $4\!\times\!4$ negation matrix} ($\alpha=1.0$,
SR\%, 20 trials). Bold diagonal = target task (the skill each negated vector aims to suppress); low diagonal SR indicates suppression. $\dagger$Task-1
anomaly (direction misalignment; see Supp.\ Sec.~G).}
\label{tab:phase1}
\small
\begin{tabular}{lrrrr}
\toprule
Model & t0 & t1 & t2 & t3 \\
\midrule
Goal (no edit)       & 100 & 100 & 100 & 95 \\
\midrule
Negate $t_0$         & \textbf{0} & 100 & 15 & 0 \\
Negate $t_1$         & 20 & \textbf{100}$^\dagger$ & 25 & 0 \\
Negate $t_2$         & 70 & 100 & \textbf{0} & 0 \\
Negate $t_3$         & 55 & 90 & 15 & \textbf{0} \\
\midrule
Random control       & 75 & 95 & 85 & 40 \\
\bottomrule
\end{tabular}
\end{table}

\begin{figure*}[t]
\centering
\includegraphics[width=\textwidth]{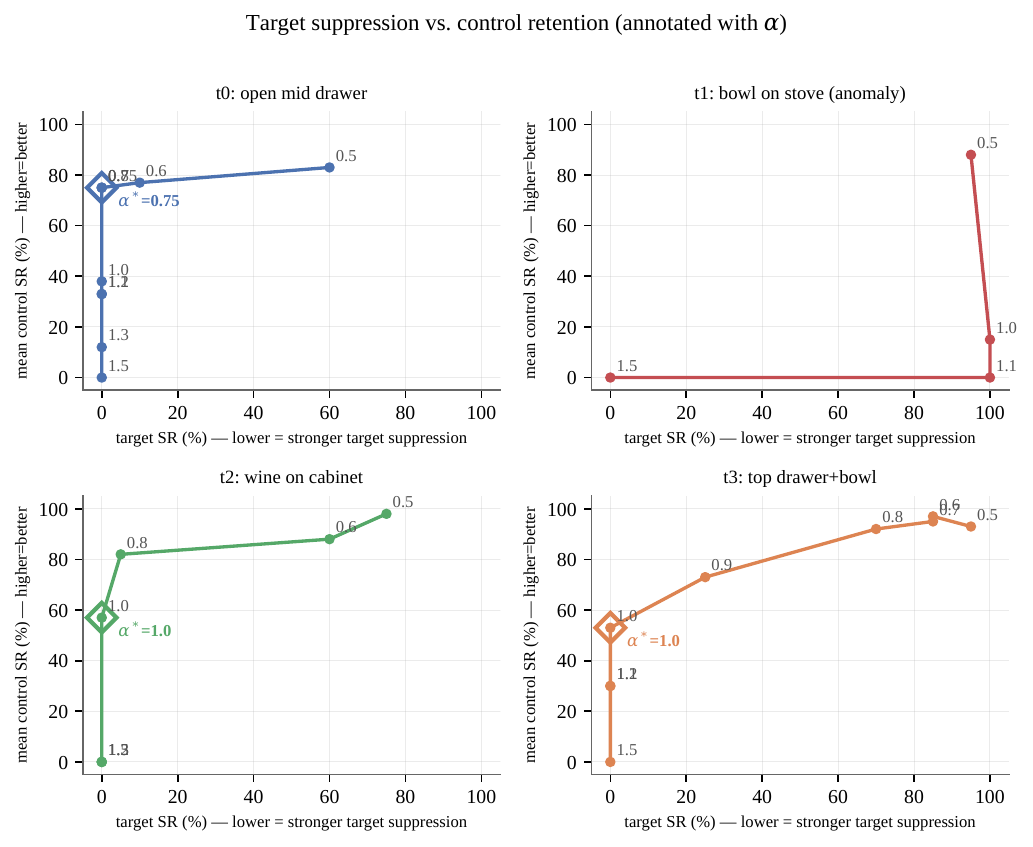}
\caption{Target suppression vs.\ control retention: target SR (lower = stronger target suppression) vs.\ mean
control SR (higher = better retention) for all four tasks.
Each point is one $\alpha$ value; diamond marks the selected constrained operating point (max control SR s.t.\ target SR${=}0$).
Task~1 is annotated with the direction-misalignment anomaly.}
\label{fig:pareto}
\end{figure*}

\begin{table}[htbp]
\centering
\caption{Cross-task effect matrix at $\alpha=1.0$.
$e_{ij}$ = relative SR drop in task~$j$ when task~$i$ is negated
($e_{ij} = 1 - \mathrm{SR}_{ij}/\mathrm{SR}^*_j$).
Diagonal entries are the target effects.}
\label{tab:effect}
\small
\begin{tabular}{lrrrr}
\toprule
& $e_{i0}$ & $e_{i1}$ & $e_{i2}$ & $e_{i3}$ \\
\midrule
Negate $t_0$ & \textbf{1.00} & 0.00 & 0.85 & 1.00 \\
Negate $t_1$ & 0.80 & \textbf{0.00}$^\dagger$ & 0.75 & 1.00 \\
Negate $t_2$ & 0.30 & 0.00 & \textbf{1.00} & 1.00 \\
Negate $t_3$ & 0.45 & 0.10 & 0.85 & \textbf{1.00} \\
\bottomrule
\end{tabular}
\end{table}

\begin{table}[htbp]
\centering
\caption{\textbf{Baseline-normalized selectivity} for all ten Goal skills
($\alpha{=}1.0$). $b_i$ = no-edit baseline; TE$_i{=}b_i{-}e_{ii}$ target
suppression; $\mathrm{Sel}_i{=}(b_i{-}e_{ii}){-}\frac{1}{N{-}1}\sum_{j\ne i}(b_j{-}e_{ij})$
difference-in-differences selectivity; CR$_i$/WCR$_i$ = mean/worst control
retention ratio (\%). $\mathrm{Sel}_i$ recapitulates the operational taxonomy: it is high
for the five selective skills, negative for the three resistant skills (target
not suppressed), and near-zero for the two global-collapse skills (where the
target is suppressed but every control is lost as well).}
\label{tab:seli}
\small
\setlength{\tabcolsep}{4.5pt}
\begin{tabular}{lrrrrr}
\toprule
Skill & $b_i$ & TE$_i$ & $\mathrm{Sel}_i$ & CR$_i$ & WCR$_i$ \\
\midrule
$t_0$ & 95  & 95  & $+40.0$ & 44 & 0 \\
$t_2$ & 100 & 100 & $+65.6$ & 64 & 0 \\
$t_3$ & 95  & 90  & $+42.2$ & 51 & 0 \\
$t_5$ & 100 & 100 & $+66.1$ & 65 & 5 \\
$t_6$ & 100 & 85  & $+50.0$ & 64 & 5 \\
\midrule
$t_1$ & 100 & 0   & $-54.4$ & 44 & 5 \\
$t_4$ & 100 & 70  & $-18.9$ & 10 & 0 \\
$t_7$ & 100 & 0   & $-61.1$ & 37 & 0 \\
\midrule
$t_8$ & 100 & 100 & $+1.7$  & 0  & 0 \\
$t_9$ & 95  & 95  & $-3.9$  & 0  & 0 \\
\bottomrule
\end{tabular}
\\[2pt]
{\footnotesize Top: five separation rows ($\mathrm{Sel}_i$ from $+40.0$ to $+66.1$). Middle: three
resistant (target not suppressed, TE small or $\mathrm{Sel}_i\!<\!0$). Bottom:
two global collapse (TE large but CR${=}0$, $\mathrm{Sel}_i\!\approx\!0$).}
\end{table}

\begin{table}[htbp]
\centering
\caption{\textbf{Separation-threshold sensitivity of the $5/3/2$ Goal taxonomy.}
Counts recomputed from the single fixed $10\!\times\!10$ matrix ($\alpha{=}1.0$) as
the separation cutoff on raw mean off-diagonal control SR is swept; the target
suppression rule ($\leq\!20\%$) is held fixed. The $5/3/2$ split is invariant for
thresholds $30$--$40\%$; only at $\geq\!45\%$ does the lowest-retention separation
row ($t_0$, raw mean control $43.9\%$) move to global collapse. The next-lowest
separation row ($t_3$) sits at $51.1\%$, so the split is in fact stable up to a
$51\%$ cutoff. The paper's $40\%$ choice is thus not a knife-edge.}
\label{tab:threshsens}
\small
\setlength{\tabcolsep}{6pt}
\begin{tabular}{cccc}
\toprule
Sep.\ threshold & Separation & Resistant & Global collapse \\
\midrule
$30\%$ & 5 & 3 & 2 \\
$35\%$ & 5 & 3 & 2 \\
$40\%$ (used) & 5 & 3 & 2 \\
$45\%$ & 4 & 3 & 3 \\
$50\%$ & 4 & 3 & 3 \\
\bottomrule
\end{tabular}
\\[2pt]
{\footnotesize Separation-row raw mean control SR: $t_0{=}43.9$, $t_3{=}51.1$,
$t_6{=}63.3$, $t_2{=}63.9$, $t_5{=}64.4$. At $\geq\!45\%$ only $t_0$ falls below the
cutoff (target still $\leq\!15\%$, so it reclassifies from separation to collapse).}
\end{table}

\begin{table}[htbp]
\centering
\caption{\textbf{Collapse-$k$}: cumulatively suppressing $k$ of the five
suppressible Goal skills at $\alpha{=}1/\sqrt{k}$ (20 rollouts/cell, mean over
tasks). Targets are suppressed, but control retention collapses at $k\!\geq\!2$:
selective suppression does not survive superposition of negation vectors.}
\label{tab:collapse_k}
\small
\setlength{\tabcolsep}{6pt}
\begin{tabular}{ccccc}
\toprule
$k$ & set $\mathcal{S}$ & $\alpha{=}1/\sqrt{k}$ & target SR & control SR \\
\midrule
1 & $\{t_0\}$              & 1.00 & \textbf{0}  & 43.9 \\
2 & $\{t_0,t_2\}$          & 0.71 & \textbf{0}  & 10 \\
3 & $\{t_0,t_2,t_3\}$      & 0.58 & \textbf{0}  & \textbf{0} \\
4 & $\{t_0,t_2,t_3,t_5\}$  & 0.50 & \textbf{0}  & \textbf{0} \\
5 & $\{t_0,t_2,t_3,t_5,t_6\}$ & 0.45 & \textbf{0}  & \textbf{0} \\
\bottomrule
\end{tabular}
\\[2pt]
{\footnotesize Each row's control SR is the mean closed-loop SR over that
edit's \emph{current} non-target skills (for $k{=}1$, the nine controls in the
$t_0$ negation row, $\GoalTZeroCtrl$; for $k\!\geq\!2$, the remaining
non-target skills). The control set therefore shrinks as $k$ grows and the
column is not a single fixed-panel metric across rows: read it as ``controls
collapse once $k\!\geq\!2$'', not as a strict monotone curve.}
\end{table}

\paragraph{Projection-based rules do not rescue multi-skill selectivity.}
A natural fix for the $k\!\geq\!2$ collapse is to project the negation vectors so
they interfere less. We tested two variants on $\{t_0,t_2\}$: \emph{retain-orth}
(each negation orthogonalized against the retained-skill subspace) and
\emph{mutual-orth} (the two negations orthogonalized against each other).
Neither escapes the collapse (Table~\ref{tab:projected}): at full strength
($\alpha{=}1.0$) both drive targets \emph{and} control $t_1$ to $0\%$, exactly
like the naive sum; at $\alpha{=}0.5$ the control is retained ($100\%$) only
because the targets are then just partially suppressed ($15$--$40\%$, not $0\%$).
Simultaneous multi-skill suppression with retained control is thus not recovered by
these projections at the tested scales.

\begin{table}[htbp]
\centering
\caption{\textbf{Projected multi-negation of $\{t_0,t_2\}$} (LIBERO-Goal, SR\,\%,
$20$ rollouts/cell). Retain-orth and mutual-orth projections both collapse the
policy at $\alpha{=}1.0$ (like the naive sum) and only retain the control at
$\alpha{=}0.5$ by leaving the targets partially suppressed.}
\label{tab:projected}
\small
\setlength{\tabcolsep}{5pt}
\begin{tabular}{llccc}
\toprule
Rule & $\alpha$ & $t_0$ & $t_2$ & ctrl $t_1$ \\
\midrule
retain-orth & $0.5$ & 25 & 40 & 100 \\
retain-orth & $1.0$ & \textbf{0} & \textbf{0} & \textbf{0} \\
mutual-orth & $0.5$ & 15 & 35 & 100 \\
mutual-orth & $1.0$ & \textbf{0} & \textbf{0} & \textbf{0} \\
\bottomrule
\end{tabular}
\end{table}

\begin{table}[htbp]
\centering
\caption{\textbf{Joint negation of $\{t_1,t_3\}$} (LIBERO-Goal, SR\,\%, $20$
rollouts/cell). Bold $=$ the two targets. At $\alpha{=}1.0$ both targets are suppressed
($0\%$), including the direction-misaligned $t_1$ that single-vector negation
leaves at $100\%$, but the controls $t_0,t_2$ \emph{also} collapse to $0\%$, so
the joint edit removes the $t_1$ target-suppression anomaly at the cost of
\emph{all} selectivity (a non-selective global collapse, like $t_8,t_9$). At
$\alpha{=}0.5$ controls are retained ($t_0{=}55$, $t_2{=}95$) only because the
targets are then not fully suppressed ($t_1{=}100$, $t_3{=}75$).}
\label{tab:multiskill13}
\small
\setlength{\tabcolsep}{5pt}
\begin{tabular}{lcccc}
\toprule
$\alpha$ & $t_0$ (ctrl) & $t_1$ (tgt) & $t_2$ (ctrl) & $t_3$ (tgt) \\
\midrule
$0.5$ & 55 & \textbf{100} & 95 & \textbf{75} \\
$1.0$ & 0  & \textbf{0}   & 0  & \textbf{0}  \\
\bottomrule
\end{tabular}
\end{table}

\begin{table}[htbp]
\centering
\caption{\textbf{Five-pair closed-loop path population} (Fig.~\ref{fig:barrier}
extended; SR\,\%, $20$ rollouts/point, anchor $\thgoal$). For each (edited
Skill~A, retained Skill~B) pair we report the negation and composition paths at
$\beta{=}1.0$, and whether the \emph{negation} path reaches a selective point
(some $\beta$ with A\,$\leq\!20\%$ and B\,$\geq\!50\%$). Negation reaches a
selective point in $2/5$ pairs (the behaviorally separable ones); composition
reaches one in \emph{none}. This is the single-pair claim of Fig.~\ref{fig:barrier}
generalized, not a universal separability guarantee.}
\label{tab:barrier_pairs}
\small
\setlength{\tabcolsep}{4pt}
\begin{tabular}{lccc}
\toprule
Pair (A/B) & neg.\ (A,B)$_{\beta=1}$ & comp.\ (A,B)$_{\beta=1}$ & neg.\ selective? \\
\midrule
$t_0/t_2$ & $(0,90)$  & $(65,0)$ & yes ($\beta{=}1$: $0,90$) \\
$t_5/t_4$ & $(0,100)$ & $(0,0)$  & yes ($\beta{=}1$: $0,100$) \\
$t_3/t_0$ & $(20,25)$ & $(0,5)$  & no \\
$t_5/t_6$ & $(0,5)$   & $(0,0)$  & no \\
$t_2/t_5$ & $(0,30)$  & $(-,0)$  & no \\
\bottomrule
\end{tabular}
\end{table}

\begin{table}[htbp]
\centering
\small
\setlength{\tabcolsep}{4pt}
\begin{tabular}{lrr rr rr rr}
\toprule
& \multicolumn{2}{c}{$t_0$} & \multicolumn{2}{c}{$t_1$} & \multicolumn{2}{c}{$t_2$} & \multicolumn{2}{c}{$t_3$}\\
\cmidrule(lr){2-3}\cmidrule(lr){4-5}\cmidrule(lr){6-7}\cmidrule(lr){8-9}
$\alpha$ & tgt & ctl & tgt & ctl & tgt & ctl & tgt & ctl\\
\midrule
0.5 & 60 & 83 & 95 & 88 & 75 & 98 & 95 & 93\\
0.6 & 10 & 77 & -- & -- & 60 & 88 & 85 & 97\\
0.7 & \textbf{0} & \textbf{75} & -- & -- & -- & -- & 85 & 95\\
0.8 & 0 & 75 & -- & -- & 5 & 82 & 70 & 92\\
0.9 & 0 & -- & -- & -- & -- & -- & 25 & 73\\
1.0 & 0 & 38 & 100 & 15 & \textbf{0} & \textbf{57} & \textbf{0} & \textbf{53}\\
1.1 & 0 & 33 & 100 & 0 & -- & -- & 0 & 30\\
1.2 & 0 & 33 & -- & -- & 0 & 0 & 0 & 30\\
1.3 & 0 & 12 & -- & -- & 0 & 0 & -- & --\\
1.5 & 0 & 0 & 0 & 0 & 0 & 0 & 0 & 0\\
\bottomrule
\end{tabular}
\caption{Measured $\alpha$-sweep (SR\%, 20 trials; ``--'' = not measured, so
this is the measured grid, not an exhaustive sweep). tgt = target-skill SR
($\downarrow$ better); ctl = mean control-skill SR ($\uparrow$ better).
Bold = best measured point in \emph{this} grid (max control SR s.t.\ target${=}0$).
The canonical $\alpha^\star$ used throughout (Tables~\ref{tab:alpha_robust},~\ref{tab:pareto}) is drawn from the coarser diagonal robustness grid: $\alpha^\star{=}0.75$ for $t_0$, which is not sampled here, so $\alpha{=}0.7$ is its nearest measured neighbor.}
\label{tab:appendix_sweep}
\end{table}

\begin{table}[htbp]
\centering
\caption{\textbf{$\alpha$-robustness of the diagonal} (target SR\,\%, 20
rollouts/cell). Suppressible skills (top) are suppressed across a broad $\alpha$ band;
resistant skills (bottom) hold until $\alpha\!\geq\!1.25$ (with $t_4$
already at $0\%$ there) and by $\alpha{=}1.5$ \emph{all} skills reach
$0\%$, collapsing non-selectively. \textbf{Bold} $=$ suppressed ($\leq\!20\%$).}
\label{tab:alpha_robust}
\small
\setlength{\tabcolsep}{5pt}
\begin{tabular}{lccccc}
\toprule
$-\tauv_i$ (diag) & $\alpha{=}0.5$ & $0.75$ & $1.0$ & $1.25$ & $1.5$ \\
\midrule
$t_0$ (suppressible) & 55 & \textbf{0} & \textbf{0} & \textbf{0} & \textbf{0} \\
$t_2$ (suppressible) & 80 & 35 & \textbf{0} & \textbf{0} & \textbf{0} \\
$t_3$ (suppressible) & 85 & 85 & \textbf{5} & \textbf{0} & \textbf{0} \\
$t_5$ (suppressible) & 95 & 100 & \textbf{0} & \textbf{0} & \textbf{0} \\
$t_6$ (suppressible) & 90 & 80 & \textbf{15} & \textbf{0} & \textbf{0} \\
\midrule
$t_1$ (aligned-lim.) & 90 & 90 & 100 & 50 & \textbf{0} \\
$t_4$ (aligned-lim.) & 100 & 100 & 30 & \textbf{0} & \textbf{0} \\
$t_7$ (aligned-lim.) & 100 & 100 & 100 & \textbf{5} & \textbf{0} \\
\bottomrule
\end{tabular}
\end{table}

\begin{table}[htbp]
\centering
\caption{Selected constrained operating point $\alpha^\star$ (smallest $\alpha$
with target SR${=}0$, highest mean control SR) per suppressible skill; see the
alpha ledger (supplement) for the full grid and selection rule.
All points achieve target SR = 0\% ($0/20$ observed, not statistically complete erasure).
$\alpha^\star$ is taken from the diagonal $\alpha$-robustness grid
(Table~\ref{tab:alpha_robust}), except $t_3$, whose $\alpha^\star{=}1.0$ comes from the
finer per-task dev sweep (Table~\ref{tab:appendix_sweep}, $t_3{=}0/20$ at $\alpha{=}1.0$);
the diagonal grid's $5\%$ there reflects rollout nondeterminism (Table~\ref{tab:multiseed}).
The dev control panel was measured for $t_0,t_2,t_3$.}
\label{tab:pareto}
\small
\begin{tabular}{lrl}
\toprule
Task & $\alpha^\star$ & Dev mean ctrl SR (\%) \\
\midrule
$t_0$ (open mid drawer)  & 0.75 & \textbf{75.0} (vs.\ 38 at $\alpha{=}1.0$) \\
$t_2$ (wine on cabinet)  & 1.0  & 56.7 \\
$t_3$ (top drawer+bowl)  & 1.0  & 53.3 \\
$t_5$ (push plate)       & 1.0  & (diagonal-grid selection) \\
$t_6$ (cream cheese)     & 1.25 & (diagonal-grid selection) \\
\bottomrule
\end{tabular}
\end{table}

\begin{table}[htbp]
\centering
\caption{\textbf{Held-out full target$\times$all-controls matrix.} Each
suppressible skill's $\alpha^\star$ is re-evaluated on the target \emph{and all
nine controls} over a \emph{disjoint} held-out set ($20$--$39$), $20$ rollouts/cell;
all ten held-out no-edit baselines were measured, so normalized retention uses a
consistent nine-control denominator (a worked normalization example is in the Code and Data Supplement).
\emph{Target suppression is robust}: all five targets stay at $0\%$ ($0/20$).
\emph{Selective control preservation is heterogeneous and fragile}: mean
baseline-normalized control retention over the nine controls ranges from $t_0$'s
$78\%$ down to $t_6$'s near-total collapse ($4\%$; mean $\HeldoutCRnorm$ across
the five rows), and four of the five rows drive at least one control to $0\%$ (worst-control
column; the $t_0$ row's worst control is $35\%$). Ctrl.\ raw $=$ mean edited control SR; Ctrl.\ norm $=$ mean of
per-control (edited\,SR $/$ no-edit baseline).}
\label{tab:heldout}
\small
\setlength{\tabcolsep}{4pt}
\begin{tabular}{llrrrr}
\toprule
Target & $\alpha^\star$ & Target SR & Ctrl.\ raw & Ctrl.\ norm & Worst ctrl. \\
\midrule
$t_0$ & $0.75$ & \textbf{0\%} & $74\%$ & $78\%$ & $35\%$ \\
$t_2$ & $1.0$  & \textbf{0\%} & $62\%$ & $65\%$ & $0\%$  \\
$t_3$ & $1.0$  & \textbf{0\%} & $47\%$ & $48\%$ & $0\%$  \\
$t_5$ & $1.0$  & \textbf{0\%} & $62\%$ & $65\%$ & $0\%$  \\
$t_6$ & $1.25$ & \textbf{0\%} & $4\%$  & $4\%$  & $0\%$  \\
\bottomrule
\end{tabular}
\end{table}

\begin{table}[htbp]
\centering
\caption{\textbf{Crosswalk: why nine-control retention reads $\GoalSepCRnorm$ in the
Introduction but $\HeldoutCRnorm$ in the held-out validation.} Both are the mean
baseline-normalized nine-control retention across the five target--control
\emph{separation} rows, measured on two different panels. \emph{Sel.\ panel} is the
fixed-$\alpha{=}1.0$ Goal $10\!\times\!10$ matrix on the selection states $0$--$19$
(the value quoted in the Introduction); \emph{Held-out} re-evaluates each row at its
\emph{selected} $\alpha^\star$ on the disjoint states $20$--$39$
(Table~\ref{tab:heldout}). The panels therefore differ in \emph{both} the state split
\emph{and} $\alpha$: the one row with a gentler $\alpha^\star\!<\!1.0$ ($t_0$) retains
\emph{more} on held-out states, while $t_6$'s $\alpha^\star{=}1.25$ turns a benign
fixed-$\alpha$ row into near-total collapse. The held-out mean ($\HeldoutCRnorm$) is
thus the more conservative, operating-point-faithful figure.}
\label{tab:crosswalk}
\small
\setlength{\tabcolsep}{6pt}
\begin{tabular}{lccc}
\toprule
Target & $\alpha^\star$ & Sel.\ CR$_{\mathrm{norm}}$ & Held-out CR$_{\mathrm{norm}}$ \\
\midrule
$t_0$ & $0.75$ & $43.9\%$ & $78\%$ \\
$t_2$ & $1.0$  & $64.3\%$ & $65\%$ \\
$t_3$ & $1.0$  & $51.5\%$ & $48\%$ \\
$t_5$ & $1.0$  & $65.4\%$ & $65\%$ \\
$t_6$ & $1.25$ & $63.9\%$ & $4\%$  \\
\midrule
Mean  & ---    & $\GoalSepCRnorm$ & $\HeldoutCRnorm$ \\
\bottomrule
\end{tabular}
\end{table}

\begin{table}[htbp]
\centering
\caption{\textbf{Direction and magnitude, not just presence, govern task-vector edits.}
MergeVLA closed-loop SR (\%, $n{=}20$/cell) on the edited task $t_0$ and a control
$t_1$. \emph{Single-vector} rows share one anchor ($\thgoal$) and one matched norm
($1.0\times$, $\lVert\tauv_0\rVert\!\approx\!83$, the full-rank reconstructed
weight-delta norm; the compact rank-$64$ LoRA-factor norm is $35.4$,
cf.\ Table~\ref{tab:rank}): flipping only the \emph{sign} of the
same vector changes the outcome, so the effect is directional, not a norm artifact.
The \emph{composition} panel varies the anchor ($\thgoal$ vs.\ the base $\thbase$) and
the aggregation (mean vs.\ sum): the outcome then depends on \emph{both} anchor and
aggregate magnitude.}
\label{tab:direction}
\small
\setlength{\tabcolsep}{4pt}
\begin{tabular}{llcc}
\toprule
& & $t_0$ (\%) & $t_1$ (\%) \\
\midrule
\multicolumn{4}{l}{\emph{Single vector} (anchor $\thgoal$, $1.0\times$ norm)} \\
\midrule
Negation & $\thgoal - \tauv_0$ & \textbf{0} & 100 \\
Addition & $\thgoal + \tauv_0$ & 50 & 90 \\
\midrule
\multicolumn{4}{l}{\emph{Composition} (anchor $\times$ aggregation)} \\
\midrule
$\thgoal$ + mean & $\tfrac12(\tauv_0{+}\tauv_1)$, $0.71\times$ & 90 & 100 \\
$\thgoal$ + sum  & $(\tauv_0{+}\tauv_1)$, $1.4\times$          & \textbf{0} & 5 \\
$\thbase$ + mean & $\tfrac12(\tauv_0{+}\tauv_1)$, $0.71\times$ & \textbf{0} & \textbf{0} \\
$\thbase$ + TIES & $\{t_0,t_1\}$, $d{=}0.3$                    & \textbf{0} & \textbf{0} \\
$\thbase$ + DARE & $\{t_0,t_1\}$, $p{=}0.9$                    & \textbf{0} & \textbf{0} \\
\bottomrule
\end{tabular}
\\[2pt]
{\footnotesize Single-vector: at matched $1.0\times$ norm, only the sign differs, yet
negation suppresses the target ($0\%$) and keeps the control ($100\%$) while addition leaves the task
half-intact ($50\%$), a directional effect. Composition: onto $\thgoal$, the smaller
mean-merge ($0.71\times$) is \emph{retained} while the larger sum ($1.4\times$)
collapses, and the same mean-merge from $\thbase$ collapses, so composition outcomes
track anchor and aggregate magnitude, not sign alone. Task vectors are near-orthogonal
($\overline{\cos}\!\approx\!0.011$ over all $90$ ordered pairs of the ten Goal skills; Supp.\ Sec.~D).}
\end{table}

\begin{table}[htbp]
\centering
\caption{\textbf{LoRA-rank sensitivity} of the task-$0$ negation ($\alpha{=}1.0$,
$20$ rollouts/cell): target $t_0$ and control $t_1$ SR (\%) across the rank used
to train the expert, with the rank-$r$ LoRA-factor norm $\lVert\tauv_0\rVert$ (the compact stored delta; the full-rank reconstructed VLM-weight-delta norm is the $\approx\!83$ used in the main text, Sec.~D). Selectivity
(control retention) is robust across rank; suppression strength is rank-dependent and
\emph{not} monotone in norm: rank-$16$ has the largest norm yet does not suppress the target.}
\label{tab:rank}
\small
\setlength{\tabcolsep}{6pt}
\begin{tabular}{lccc}
\toprule
LoRA rank & $\lVert\tauv_0\rVert$ & target $t_0$ & control $t_1$ \\
\midrule
$16$ & $60.0$ & $95$ & $95$ \\
$32$ & $40.6$ & \textbf{5}  & $100$ \\
$64$ & $35.4$ & \textbf{0}  & $100$ \\
\bottomrule
\end{tabular}
\end{table}

\paragraph{Training-seed robustness probe ($t_0$/$t_1$).} The full negation
matrices in this paper use a single expert-training seed. As a targeted
robustness check, \emph{not} a full multi-seed replication of every
matrix, we independently re-trained the task-$0$ expert two more times ($N{=}3$
runs, each configured for $10$k steps, identical config and shared init, differing
only in the nondeterministic data-shuffle; the third run's step-$10000$ checkpoint
is the one merged and negated, though only its step-$4000$ stdout survives; the
checkpoint manifest and per-episode outcomes are in the Code and Data Supplement)
and negated $\tauv_0$, scoring target $t_0$ and
control $t_1$ ($20$ rollouts/cell; goal-checkpoint baseline
$t_0/t_1{=}95\%/100\%$). Evaluated at a \emph{fixed} $\alpha{=}1.0$, all three seeds
show a positive target--control gap (the target stays below its control), but the
target is fully suppressed only for seed~1 and near-suppressed for seed~3, while seed~2
stays at $70\%$ (Table~\ref{tab:multiseed}, left); the \emph{suppression threshold}
$\alpha$ is thus run-dependent. For seed~3 we ran a full per-seed $\alpha$ sweep
(Table~\ref{tab:multiseed}, right): it is selective at $\alpha\!=\!1.0$--$1.25$
(target $\leq\!5\%$, control $\geq\!85\%$) and collapses non-selectively only at
$\alpha\!=\!1.5$, the same dose--response shape as the main run. Thus the
\emph{sign} of the target--control gap is reproduced across all three seeds, while
the suppression threshold is run-dependent. Evaluating each seed's $t_0$ negation ($\alpha{=}1.0$) against a
\emph{fixed} five-task control panel ($t_1$--$t_5$) gives mean control retention
of $45\%/60\%/73\%$ for seeds $1/2/3$; controls are broadly retained but
heterogeneous across both seed and control task. This probes one target skill,
not the full matrix, so we do not claim the single-seed limitation is removed
for the matrices.

\begin{table}[htbp]
\centering
\caption{\textbf{Training-seed robustness probe} for $\tauv_0$ negation
($20$ rollouts/cell; cells are success rates, e.g.\ a $0/20$ cell has $95\%$
Wilson interval $[0,16]\%$). \emph{Left:} three independent $10$k-step seeds
evaluated at a \emph{fixed} $\alpha{=}1.0$: the target stays below its control in
all three (positive target--control gap), though at $\alpha{=}1.0$ the target is
fully suppressed only for seed~1 ($0\%$) and near-suppressed for seed~3 ($5\%$), while
seed~2 remains at $70\%$; the \emph{suppression threshold} is run-dependent.
\emph{Right:} seed-3 per-seed $\alpha$ sweep: selective through $\alpha{=}1.25$,
non-selective collapse at $1.5$. Goal $t_0/t_1$ baseline $95\%/100\%$.}
\label{tab:multiseed}
\small
\setlength{\tabcolsep}{4pt}
\begin{tabular}{lcc@{\hspace{1.5em}}lcc}
\toprule
\multicolumn{3}{c}{Three $10$k seeds} & \multicolumn{3}{c}{Seed-3 $\alpha$ sweep} \\
\cmidrule(r){1-3}\cmidrule(l){4-6}
Seed ($\alpha$) & $t_0$ & $t_1$ & $\alpha$ & $t_0$ & $t_1$ \\
\midrule
1 ($1.0$) & $0$   & $100$ & $0.5$  & $85$ & $95$ \\
2 ($1.0$) & $70$  & $95$  & $0.75$ & $90$ & $100$ \\
3 ($1.0$) & $5$   & $85$  & $1.0$  & $5$  & $85$ \\
          &       &       & $1.25$ & $0$  & $85$ \\
          &       &       & $1.5$  & $0$  & $0$ \\
\bottomrule
\end{tabular}
\\[2pt]
{\footnotesize Canonical seed-3 $\alpha{=}1.0$ target is $5\%$ ($1/20$, from the
$\alpha$-sweep run). An earlier $20$-trial eval of the \emph{same} step-$10000$
checkpoint on the \emph{same} fixed initial states gave $20\%$ ($4/20$); its
control also differed ($80\%$ vs.\ $85\%$), so the difference is closed-loop
rollout nondeterminism (MuJoCo dynamics and bf16 forward passes are not
bit-reproducible), not a protocol or panel change. The target stays well below
its control either way. We report the $\alpha$-sweep value as canonical.}
\end{table}

\begin{figure*}[htbp]
\centering
\includegraphics[width=\textwidth]{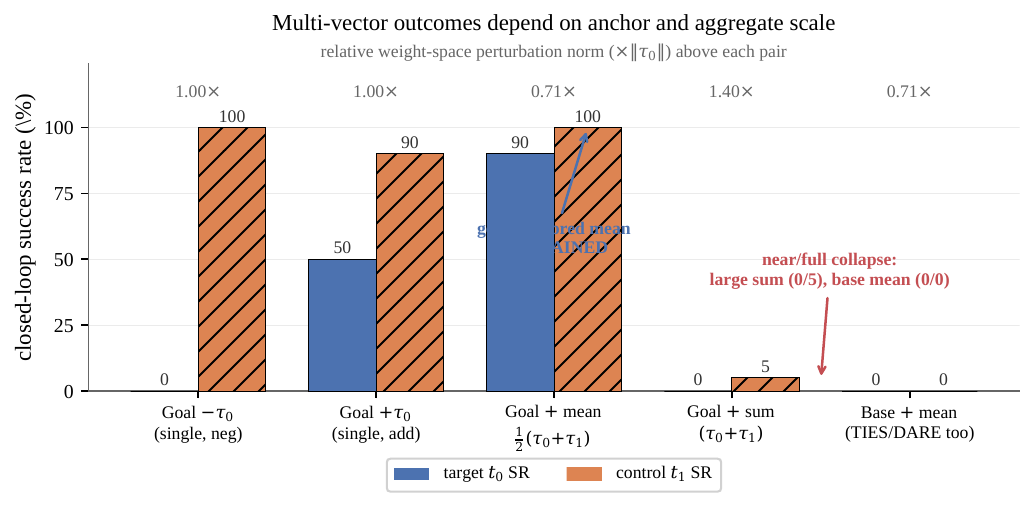}
\caption{\textbf{Multi-vector outcomes depend on anchor and aggregate scale.}
Target $t_0$ (solid) vs.\ control $t_1$ (hatched) closed-loop SR for each edit
($20$ trials/cell), with the relative weight-space perturbation norm annotated
above. Sign is decisive for a single vector at matched norm: goal-anchored
negation is selective (target $0\%$, control $100\%$) while goal-anchored
addition keeps the target ($50\%$). For \emph{composition}, the outcome tracks
the anchor and aggregate magnitude, not merely ``addition'': the goal-anchored
\emph{mean} ($0.71\times$) is \emph{retained} ($90\%/100\%$), whereas the
goal-anchored \emph{sum} ($1.4\times$) and base-anchored merges (mean, TIES,
DARE) collapse both skills. Magnitude alone does not explain the outcome: the
retained goal-mean and the collapsing base-mean share the same $0.71\times$ norm.}
\label{fig:direction}
\end{figure*}

\section{C. Generality and Boundary: Cross-Suite and Cross-Architecture}
\label{asec:generality}
To test generality and locate where the effect ends, we compute the \emph{complete} $4\!\times\!4$ negation matrix on the two remaining LIBERO suites and a second backbone. Table~\ref{tab:spatial} gives the full LIBERO-Spatial matrix (every negated $\thgoal-\tauv_i$ scored on all four tasks, $\alpha{=}1.0$). Spatial retains higher average off-diagonal utility ($61.7\%$ over all $12$ off-diagonal cells) than Object/Long10, but this aggregate is inflated by the two resistant targets whose edits leave controls high while failing to suppress their own targets: its diagonal is heterogeneous: $t_0$ is suppressed ($0\%$), $t_2$ partially ($70\%$), and $t_1/t_3$ resist ($90\%/100\%$); and the one clean suppression ($t_0$) also drops its controls, so \emph{no row is cleanly selective} ($0/4$). We therefore read Spatial as \emph{intermediate}: target suppression occurs but never with controls retained, more separable than the entangled suites, yet with none of Goal's clean separation rows.

The boundary appears on the entangled suites. Table~\ref{tab:object} shows that LIBERO-Object skills share a single motor primitive (pick-place-into-basket over disjoint objects), so negation still \emph{fires}: $3/4$ targets fall to $\leq\!10\%$, with $t_0$ a resistant outlier, but is \emph{not} selective, as the controls collapse alongside the targets. Table~\ref{tab:long10} tells the same story for the long-horizon compound tasks: shared sub-goals mean only $2/4$ targets are suppressed and controls collapse regardless, despite these task vectors being as near-orthogonal as Goal's (main paper). This indicates that separability, not weight-space orthogonality, governs whether suppression stays local.

Finally, Table~\ref{tab:openvla} reproduces the full LIBERO-Goal negation matrix on OpenVLA-7B ($10$ rollouts/cell, $\alpha{=}1.0$), a distinct backbone. Only three targets have a diagnostic no-edit baseline ($t_0,t_1,t_2$ at $70/90/70\%$); $t_3$ has a $10\%$ no-edit baseline and is \emph{non-diagnostic}, so it does not enter the suppression count. The three high-baseline diagnostic targets are suppressed to $0\%$ after negation, while off-diagonal controls average $50.8\%$. This target suppression on the three diagnostic targets supports transfer of the qualitative target--control pattern; we rest the claim on that structure rather than the off-diagonal point estimates, since the $10$-trial cells leave absolute margins backbone-dependent.

\begin{table}[htbp]
\centering
\caption{\textbf{LIBERO-Spatial full $4\!\times\!4$ negation matrix} (SR\%, 20
trials/cell, $\alpha{=}1.0$). $\thgoal=$ MergeVLA-Spatial; row $=$ negated
$\tauv_i$, column $=$ evaluated task; \textbf{bold} diagonal $=$ target.
Off-diagonal mean $61.7\%$, but inflated by the resistant rows.
Spatial is \emph{intermediate} with $0/4$ clean selective rows: $t_0$'s target
is suppressed but its own controls also drop to $0\%$, $t_2$ is only partially
suppressed, and $t_1/t_3$ resist; no single row shows
target-suppressed-with-controls-retained, so Spatial is not as clean as Goal.}
\label{tab:spatial}
\small
\setlength{\tabcolsep}{6pt}
\begin{tabular}{lcccc}
\toprule
$-\tauv_i\!\downarrow$ & $t_0$ & $t_1$ & $t_2$ & $t_3$ \\
\midrule
$-\tauv_0$ & \textbf{0}   & 0   & 0    & 0   \\
$-\tauv_1$ & 95  & \textbf{90}  & 100  & 100 \\
$-\tauv_2$ & 15  & 40  & \textbf{70}   & 90  \\
$-\tauv_3$ & 100 & 100 & 100  & \textbf{100} \\
\bottomrule
\end{tabular}
\\[2pt]
{\footnotesize Goal-model baseline $t_0/t_1/t_2/t_3=95/100/100/100$. Diagonal
(target): $t_0$ suppressed ($0$), $t_2$ partial ($70$), $t_1/t_3$ resist
($90/100$). $95\%$ Wilson for a $0/20$ cell is $[0,16]\%$.}
\end{table}

\begin{table}[htbp]
\centering
\caption{\textbf{LIBERO-Object full $4\!\times\!4$ negation matrix} (SR\,\%, 20
rollouts/cell, $\alpha{=}1.0$). Row $=$ negated $\tauv_i$, column $=$ evaluated
task; \textbf{bold} diagonal $=$ target. Object skills share one motor
primitive (pick-place-into-basket, disjoint objects), so suppression fires ($3/4$
targets $\leq\!10\%$; $t_0$ a resistant outlier) but is \emph{not}
selective: off-diagonal mean is only $5.8\%$ ($12$ cells), i.e.\ controls
collapse with the targets.}
\label{tab:object}
\small
\setlength{\tabcolsep}{6pt}
\begin{tabular}{lcccc}
\toprule
$-\tauv_i\!\downarrow$ & $t_0$ & $t_1$ & $t_2$ & $t_3$ \\
\midrule
$-\tauv_0$ & \textbf{85} & 15 & 10 & 15 \\
$-\tauv_1$ & 0  & \textbf{0}  & 0  & 0  \\
$-\tauv_2$ & 0  & 0  & \textbf{0}  & 0  \\
$-\tauv_3$ & 5  & 5  & 20 & \textbf{10} \\
\bottomrule
\end{tabular}
\\[2pt]
{\footnotesize Goal-model baseline $t_0/t_1/t_2/t_3=95/100/100/100$. Suppression
succeeds on $3/4$ targets but selectivity does not (off-diagonal mean $5.8\%$),
placing Object between Goal/Spatial and Long10.}
\end{table}

\begin{table}[htbp]
\centering
\caption{\textbf{LIBERO-Long10 full $4\!\times\!4$ negation matrix} (SR\,\%, 20
rollouts/cell, $\alpha{=}1.0$). Row $=$ negated $\tauv_i$, column $=$ evaluated
task; \textbf{bold} diagonal $=$ target. Long-horizon compound skills reuse
shared sub-goals, so negation is \emph{non-selective}: only $2/4$ targets are suppressed
and controls collapse regardless: off-diagonal mean $11.7\%$ ($12$ cells).}
\label{tab:long10}
\small
\setlength{\tabcolsep}{6pt}
\begin{tabular}{lcccc}
\toprule
$-\tauv_i\!\downarrow$ & $t_0$ & $t_1$ & $t_2$ & $t_3$ \\
\midrule
$-\tauv_0$ & \textbf{0}  & 0  & 0  & 0  \\
$-\tauv_1$ & 0  & \textbf{15} & 15 & 0  \\
$-\tauv_2$ & 0  & 0  & \textbf{60} & 25 \\
$-\tauv_3$ & 0  & 10 & 90 & \textbf{55} \\
\bottomrule
\end{tabular}
\\[2pt]
{\footnotesize Goal-model baseline $t_0/t_1/t_2/t_3=75/100/100/100$. Only
$t_0,t_1$ are suppressed; off-diagonal mean $11.7\%$, so selectivity, not just
suppression, is what fails.}
\end{table}

\begin{table}[htbp]
\centering
\caption{\textbf{Target suppression reproduces on OpenVLA-7B (three diagnostic targets).}
Closed-loop success rate (\%, 10 rollouts/cell) on LIBERO-Goal after
negating each task vector ($\thedit=\thgoal-\alpha\,\tauv_i$,
$\alpha=1.0$); rows are the negated skill, columns the evaluated skill.
\textbf{Diagonal} (target) cells mark the skill each vector aims to suppress.
Only $t_0,t_1,t_2$ have a diagnostic no-edit baseline ($70/90/70\%$); $t_3$'s
no-edit baseline is $10\%$, so it is \emph{non-diagnostic} and its diagonal
does not enter the suppression count. The three diagnostic targets are
suppressed to $0\%$, while off-diagonal control skills are largely preserved
(mean $50.8\%$). Target suppression thus transfers to a backbone differing from
MergeVLA in scale ($\sim$10$\times$), action representation (discrete tokens
vs.\ continuous head), and pretraining (Open-X).}
\label{tab:openvla}
\small
\begin{tabular}{lcccc}
\toprule
 & \multicolumn{4}{c}{Evaluated skill (SR\,\%)} \\
\cmidrule(lr){2-5}
Edit $\downarrow$ & $t_0$ & $t_1$ & $t_2$ & $t_3$ \\
\midrule
Goal baseline (no edit) & 70 & 90 & 70 & 10 \\
\midrule
Negate $\tauv_0$ & \textbf{0} & 100 & 80 & 30 \\
Negate $\tauv_1$ & 60 & \textbf{0} & 80 & 30 \\
Negate $\tauv_2$ & 30 & 60 & \textbf{0} & 40 \\
Negate $\tauv_3$ & 0 & 40 & 60 & \textbf{0} \\
\midrule
Compose $\sum_i\tauv_i$ & 0 & 0 & 0 & 0 \\
\bottomrule
\end{tabular}
\\[2pt]
{\footnotesize Bold = negation target (diagonal). The goal model is itself
weak on $t_3$ ($10\%$ no-edit), so $t_3$ is non-diagnostic and excluded from
the suppression count; the transfer claim rests on the three high-baseline
diagnostic targets ($t_0,t_1,t_2$), all suppressed to $0\%$, with mean
off-diagonal $50.8\%$. Composition ($\thgoal+\alpha\sum_i\tauv_i$,
$\alpha{=}1.0$) collapses \emph{every} skill to $0\%$, mirroring MergeVLA.
Off-diagonal cells use 10 rollouts; the diagonal (target) cells were
re-evaluated at 20 rollouts and remain $0/20$, tightening the $95\%$ Wilson
upper bound on the suppressed-target claim from $28\%$ to $16\%$. (95\% Wilson
interval: a $0/20$ cell is $[0,16]\%$; a $0/10$ cell $[0,28]\%$; $80\%$ at
$n{=}10$ is $[49,94]\%$.)}
\end{table}

\section{D. Diagnostics: Sharpness, Geometry, and Localization}
\label{asec:diagnostics}
\paragraph{Loss-landscape sharpness and its confounds.} Tables~\ref{tab:sharpness} and~\ref{tab:sharpness_v2}, together with Figure~\ref{fig:sharpness}, establish a consistent geometric signature: the $\varepsilon$-sharpness ordering runs Expert $\gg$ Goal $\gg$ Negated, with negated models the flattest of the three (negated $\sim$2.4$\times$ vs.\ experts' $\sim$15--17$\times$). Because the naive sharpness ratio divides by base loss, and negated models carry a higher base loss, this ordering could in principle be an artifact of the denominator. Table~\ref{tab:sharpness_v2} discharges this worry three ways: the raw \emph{absolute} maximum-over-$20$-sampled-perturbations increase $\max_n[L(\theta{+}\delta_n){-}L(\theta)]$ (no denominator) preserves the ordering, as do the scale-normalized ASAM and filter-normalized metrics, and it survives a weight-norm control. We stress that this evidence is \emph{correlational}: sharpness co-varies with suppression state but the sharpness measurements alone do not license a causal claim; the additional weight is carried by the closed-loop path behavior and the norm-matched SAM intervention reported below (Supp.\ Sec.~G). (Our sharpness is a maximum over $20$ sampled Gaussian perturbations at a fixed mini-batch, not a true worst-case local maximization.)

\paragraph{Geometry does not predict control survival; no single depth band suffices at its native norm.} Figure~\ref{fig:geometry} shows that skill directions live overwhelmingly in the VLM weights, which are near-orthogonal ($\overline{\cos}\approx0.014$ off-diagonal on the four-skill probe), whereas the action-head vectors are far more aligned ($\cos\approx0.26$). To rule out that this near-orthogonality is a small-sample artifact of the four-skill probe, we recompute the geometry over the \emph{full} set of ten LIBERO-Goal skill vectors (all $90$ ordered pairs, over the complete $1.25\!\times\!10^{9}$-parameter VLM weight space; per-tensor deltas in \texttt{float32}, accumulated in \texttt{float64}): the off-diagonal cosine is tightly concentrated near zero (mean $0.011$, median $0.011$, range $[0.003,0.018]$, s.d.\ $0.003$), with VLM-only reconstructed skill-vector norms in a narrow band ($71.2$--$89.7$; these are the full-rank VLM-weight deltas, larger than the rank-64 LoRA-factor norm of $\approx\!35$ reported in Supp.\ Table~\ref{tab:rank}). The distribution thus confirms uniform near-orthogonality across every skill pair rather than a cherry-picked corner. One might therefore expect near-orthogonality to guarantee selective suppression, but the main-paper geometry--selectivity figure refutes this cleanly: the mean pairwise VLM cosine is near-identical ($\approx0.014$) across all four suites, yet control-skill survival after single-skill negation spans an order of magnitude. Near-orthogonality is thus a descriptive geometric property, not a predictor of control survival. Table~\ref{tab:layerband} localizes the edit along depth: restricting $\tauv_i$ to any single band of the 24-layer LM stack leaves the target intact ($80$--$100\%$), and only the full-depth vector suppresses it ($0\%$). Because each single band carries a smaller-norm perturbation than the full vector, this shows the full-depth vector is required \emph{at the tested per-band scale}, rather than proving skill identity is distributed across depth; norm-matched band ablations are future work. The component ablation in Table~\ref{tab:ablation} complements this, isolating which submodules of the task-0 edit drive target collapse versus control retention.

\paragraph{Collateral interference has structure but no cheap predictor.} The interference graph over the $8\!\times\!8$ Goal matrix is shown in the main paper (Fig.~\ref{fig:collateral}A): the five suppressible skills form a densely interfering cluster ($14$ edges $\geq\!50$\,pp, with $t_3$ the highest-degree collateral hub), while the three resistant skills ($t_1,t_4,t_7$) carry no edges, not because their edits are harmless, but because the edge rule draws collateral only from source rows whose own target is suppressed, and as controls these three are rarely driven down $\geq\!50$\,pp by a suppressible edit (negating a resistant vector instead damages controls only through non-selective norm collapse, Supp.\ Table~\ref{tab:seli}). Yet this per-cell structure resists prediction from any cheap non-circular feature: across all $56$ off-diagonal cells, neither the pairwise weight-space cosine $\lvert\cos(\tauv_i,\tauv_j)\rvert$ nor a finite-difference cross-task loss sensitivity correlates with the observed collateral drop (Spearman $\rho\!=\!-0.03$ and $-0.07$). Because the $56$ cells are not independent (each shares a negated-row model and an evaluated-task column), we assess significance with a negated-row (cluster) permutation test rather than an IID $p$-value: the observed $\rho$ sits well inside the permuted null (permutation $p\!>\!0.6$), and the result is stable under leave-one-row-out. This is a deliberately reported negative: consistent with the cross-suite result above, collateral damage is a closed-loop behavioral quantity that weight geometry describes but does not forecast.

\paragraph{Negation is a reversible weight-space toggle (float32 round-trip).} Complementing the behavioral masking result, the edit is algebraically invertible: reapplying the same vector, $(\thgoal-\alpha\tauv_i)+\alpha\tauv_i$, recovers $\thgoal$. \emph{When the intermediate is kept in} \texttt{float32}, the round-trip is exact: $\max_k\lvert\Delta_k\rvert=0$ across all $1.25\!\times\!10^{9}$ parameters, so the restored checkpoint is bit-identical to the goal model and its behavior is identical. This bit-exactness is a property of the \texttt{float32} implementation, not a precision-free guarantee: the shipped policy is served and evaluated in \texttt{bfloat16}, and a \texttt{bfloat16} round-trip perturbs ${\sim}8.5\%$ of weights by $\leq\!10^{-3}$; we therefore report reversibility as an exact \texttt{float32} round-trip with a bounded, small \texttt{bfloat16} serialization error, not an unconditional algebraic identity at inference precision.

\paragraph{Method positioning and the suppression frontier.} Table~\ref{tab:comparison} situates negation among unlearning approaches: like other task-arithmetic edits it is ``instant'' (a single closed-form edit) and, uniquely among the rows, data-free \emph{and} gradient-free at edit time for a VLA. Table~\ref{tab:ga} reports the full matched-panel E1 comparison against the gradient baselines and the leave-one-task-out retraining oracle: the same target $t_0$, the same nine controls $t_1$--$t_9$, the same held-out initial states, $n{=}20$/cell. All five removal methods suppress the target ($0/20$). Plain gradient ascent collapses \emph{every} control ($0\%$ mean); the retain-aware methods are \emph{not} uniformly better than the data-free edit: Gradient-Difference ($88.3\%$ raw mean control) exceeds negation ($73.9\%$), whereas NegGrad+ ($66.1\%$) does not and drives control $t_9$ to $0\%$; the retraining oracle is the upper bound ($95.6\%$) but costs a full retrain. Per-control counts and per-method aggregates are in Table~\ref{tab:ga}. Negation's niche is thus the removal-time data- and gradient-free regime (given precomputed expert vectors; not training-free overall, as the per-skill expert bank is trained upstream, Table~\ref{tab:comparison}) on this cost--retention frontier, not universal dominance.

\paragraph{Relearning: faster recovery is consistent with masking.} Table~\ref{tab:relearn} probes what negation does to the weights by re-finetuning the suppressed skill $t_0$ from three checkpoints: negated, base (never-trained-on-$t_0$ task-naive floor), and goal (intact ceiling). The negated policy recovers far faster than the task-naive floor, \emph{consistent with} the skill's substrate being preserved and behaviorally masked rather than erased at the weight level. This mirrors the LLM-unlearning literature, where a small amount of relearning or benign fine-tuning recovers capabilities that behavioral unlearning appeared to remove, indicating suppression of access rather than deletion from the weights~\cite{lynch2024eight,deeb2024unlearning,lucki2024adversarial}. This is a single-skill, single-seed observation (a caveat for any safety use) and does not by itself exclude all forms of weight-level forgetting.

\paragraph{Representational corroboration (CKA).} Table~\ref{tab:cka} reports linear centered kernel alignment (CKA) between the intact goal model and the negated ($-\tauv_0$) model, computed on hidden states over skill-$t_0$ inputs, mean-pooled per layer across the $25$-layer hidden-state stack (the $24$ LM blocks plus the embedding output). CKA stays high at every depth (overall mean $0.89$, range $0.84$--$0.93$), i.e.\ the negated model's representations on the suppressed task remain close to the intact model's, consistent with information being latent (masked) rather than removed. We note the scope: this is a single edited skill ($t_0$), single seed, linear CKA with mean-pooling; high CKA is corroborating, not by itself proof of recoverability (the relearning curve provides the behavioral evidence).

\begin{table}[htbp]
\centering
\caption{\textbf{Layerwise linear CKA} between intact and negated ($-\tauv_0$)
hidden states on suppressed-task $t_0$ inputs (mean-pooled per layer, $25$-layer
stack). CKA stays high at all depths ($0.84$--$0.93$), consistent with
representations being largely preserved (masking). \emph{Caveats}: single skill,
single seed, linear CKA, mean-pooled over the rollout; we did not record a
matched sample count or a shuffled-pair null baseline, so we read this as
suggestive corroboration of the relearning result, not a standalone claim.}
\label{tab:cka}
\small
\setlength{\tabcolsep}{6pt}
\begin{tabular}{lccc}
\toprule
Layer band & mean CKA & min & max \\
\midrule
Early ($0$--$7$)   & $0.887$ & $0.865$ & $0.907$ \\
Mid ($8$--$16$)    & $0.896$ & $0.880$ & $0.932$ \\
Late ($17$--$24$)  & $0.888$ & $0.840$ & $0.910$ \\
\midrule
All ($0$--$24$)    & $0.890$ & $0.840$ & $0.932$ \\
\bottomrule
\end{tabular}
\end{table}

\begin{table}[htbp]
\centering
\caption{$\varepsilon$-sharpness and base loss by model type.}
\label{tab:sharpness}
\small
\begin{tabular}{llrr}
\toprule
Model & Eval data & Sharpness & Base loss \\
\midrule
Expert ($t_0$)  & task 0 & 14.9$\times$ & 0.034 \\
Expert ($t_1$)  & task 1 & 17.7$\times$ & 0.030 \\
\midrule
Goal (no edit)  & task 0 & 7.0$\times$  & 0.067 \\
Goal (no edit)  & task 1 & 4.1$\times$  & 0.084 \\
\midrule
Negate $t_0$, $\alpha$=1 & task 0 (target)  & 3.3$\times$ & 0.130 \\
Negate $t_0$, $\alpha$=1 & task 1 (pres.)   & 3.2$\times$ & 0.140 \\
Negate $t_2$, $\alpha$=1 & task 2 (target)  & 2.5$\times$ & 0.158 \\
Negate $t_3$, $\alpha$=1 & task 3 (target)  & 2.6$\times$ & 0.152 \\
\bottomrule
\end{tabular}
\end{table}

\begin{table}[htbp]
\centering
\caption{Sharpness ordering Expert $\gg$ Goal $\gg$ Negated is robust to
three confounds. \textbf{Absolute} is the raw worst-case loss increase
$\max_n[L(\theta{+}\delta_n){-}L(\theta)]$ (\emph{no} base-loss denominator,
so it cannot be an artifact of negated models' higher base loss); \textbf{ASAM}
and \textbf{Filter-norm} are reparameterization-invariant relative metrics.
All model types share near-identical weight norm ($\|\theta\|\!\approx\!1937$,
within $0.5\%$), so the ordering is not a magnitude artifact either. Means
over model types, $\varepsilon\!=\!0.01$.}
\label{tab:sharpness_v2}
\small
\begin{tabular}{lrrr}
\toprule
Model type & Absolute ($\Delta L$) & ASAM (rel.) & Filter-norm \\
\midrule
Expert  & $285$ & 15.1$\times$ & 17.1$\times$ \\
Goal    & $92$  & 4.3$\times$  & 4.1$\times$ \\
Negated & $58$  & \textbf{2.4$\times$}  & \textbf{2.4$\times$} \\
\bottomrule
\end{tabular}
\end{table}

\begin{figure*}[htbp]
\centering
\includegraphics[width=\textwidth]{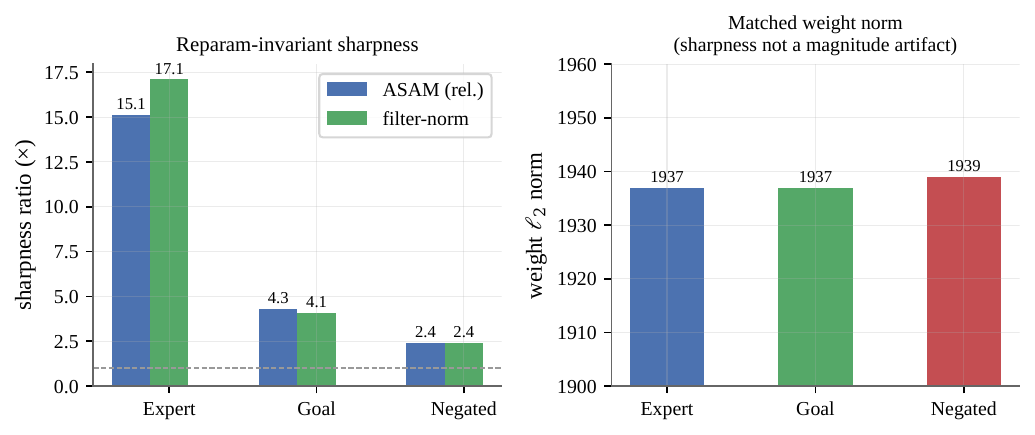}
\caption{\textbf{The sharpness ordering Expert $\gg$ Goal $\gg$ Negated
is a geometric property, robust to three confounds.}
Left: \emph{sampled perturbation-sensitivity} ratio (max over $20$ Gaussian
perturbations at a fixed mini-batch, defined in \S\,A: a sampled maximum, not a true
worst-case $\varepsilon$-sharpness) under two reparameterization-invariant
metrics (ASAM, filter-normalized); negated models are flattest
($\sim$2.4$\times$) while experts are sharpest ($\sim$15--17$\times$).
Right: all three model types share near-identical weight $\ell_2$ norm
($\approx$1937), ruling out the Dinh et al.\ scale artifact. The same
ordering holds for the \emph{absolute} loss increase ($285\!\gg\!92\!\gg\!58$,
no base-loss denominator; Table~\ref{tab:sharpness_v2}), so it is not an
artifact of negated models' higher base loss either.}
\label{fig:sharpness}
\end{figure*}

\begin{figure*}[htbp]
\centering
\includegraphics[width=\textwidth]{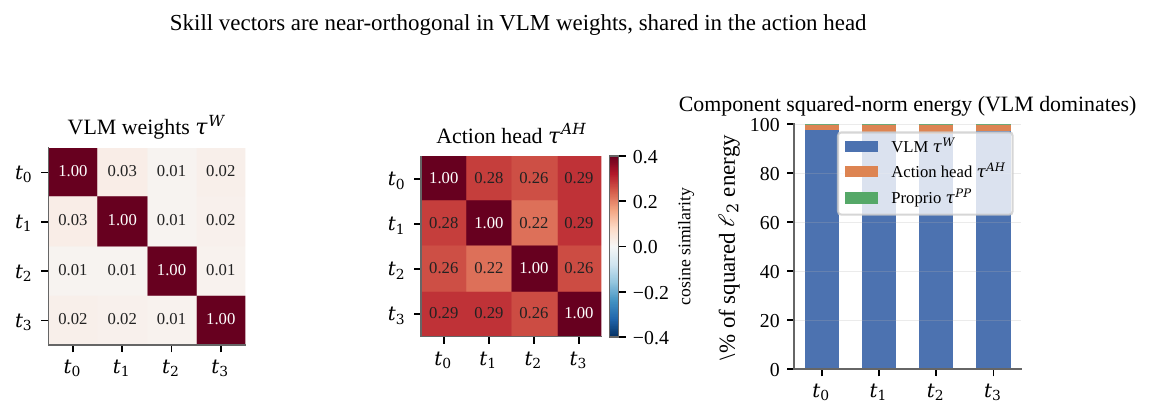}
\caption{\textbf{Skill-vector geometry.} \emph{Left, middle:} cosine
similarity between task vectors. VLM weights $\tauv^W$ are near-orthogonal
($\overline{\cos}\approx0.014$ off-diagonal, four-skill probe; $0.011$ over all
$90$ ordered pairs of the ten Goal skills), so each skill occupies a distinct
direction; action-head vectors $\tauv^{\mathrm{AH}}$ are far more aligned
($\cos\approx0.26$), reflecting shared low-level motor primitives.
\emph{Right:} per-task component decomposition of the squared $\ell_2$ energy
(squared norms are additive, so shares sum to $100\%$): the VLM component carries
$\approx\!97\%$ of each skill vector's squared $\ell_2$ energy, equivalently
$\lVert\tauv^W\rVert/\lVert\tauv\rVert\!\approx\!0.98$, so the task-vector energy and
the tested suppression effect concentrate in the VLM component (Supp.\ Sec.~H).}
\label{fig:geometry}
\end{figure*}

\begin{table}[htbp]
\centering
\caption{\textbf{Cross-architecture, cross-suite summary.} Under single-skill
negation ($\alpha{=}1.0$, single training seed). ``Selective'' counts skills
whose target is suppressed \emph{and} controls retained; mean control SR is the
raw off-diagonal mean over each suite's \emph{full} negation matrix (Goal
$10\!\times\!10$, others $4\!\times\!4$). On MergeVLA-Goal the ten skills split $5$ selective /
$3$ resistant / $2$ non-selective global collapse. Clean selective rows appear
only on MergeVLA-Goal, OpenVLA-Goal, and \piZeroFive{}-Goal; on Spatial the
target is suppressed but \emph{no} row is cleanly selective ($0/4$; its high
off-diagonal is inflated by resistant targets), and on Object/Long10/\piZeroFive{}-Long
the target is suppressed but controls collapse with it. Rollouts/cell: $20$ (MergeVLA and
\piZeroFive{}-Goal), $10$ (OpenVLA off-diagonal and \piZeroFive{}-Long; OpenVLA
diagonals re-run at $20$). This tally consolidates Fig.~\ref{fig:collateral}C.}
\label{tab:summary}
\small
\setlength{\tabcolsep}{3pt}
\begin{tabular}{llccc}
\toprule
Arch. & Suite & Sep.\ rows & Mean ctrl SR & Any sep.\ row? \\
\midrule
MergeVLA & Goal    & $5/10$\rlap{$^\ast$} & $37.7\%$ & yes \\
MergeVLA & Spatial & $0/4$ & $61.7\%$\rlap{$^\ddagger$} & no \\
MergeVLA & Object  & $0/4$ & $5.8\%$  & no \\
MergeVLA & Long10  & $0/4$ & $11.7\%$ & no \\
OpenVLA  & Goal    & $3/3$\rlap{$^\dagger$} & $50.8\%$ & yes \\
\piZeroFive{} & Goal & $3/4$\rlap{$^\S$} & $54.6\%$ & yes \\
\piZeroFive{} & Long & $0/4$ & $3.3\%$  & no \\
\bottomrule
\end{tabular}
\\[2pt]
{\footnotesize $^\ast$$3$ resistant (target not suppressed), $2$
target-suppressed but non-selective. $^\dagger$OpenVLA $t_3$ (baseline $10\%$)
excluded as non-diagnostic; $3$ high-baseline targets counted.
$^\ddagger$Spatial's high off-diagonal is inflated by two resistant targets
($t_1,t_3$) whose edits leave controls high while failing to suppress their own targets; no row is cleanly selective.
$^\S$\piZeroFive{}-Goal: all four rows have a positive target--control gap
($30.0$--$68.3$\,pp), but row $s_3$ ($=t_2$) has mean control $30\%$ ($<\!40\%$),
so it is a positive-gap but non-selective row; $s_0,s_1,s_2$ are selective.}
\end{table}

\begin{figure*}[htbp]
\centering
\includegraphics[width=\textwidth]{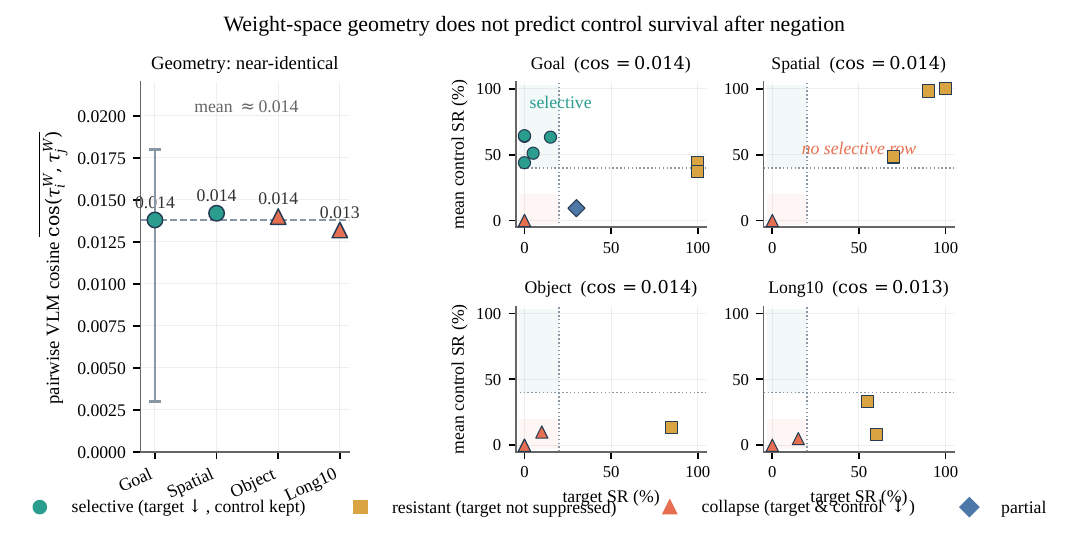}
\caption{\textbf{Per-suite target-vs-control facets (expanded companion to
Fig.~\ref{fig:collateral}B--C).} \emph{Left:} mean pairwise VLM task-vector
cosine is near-identical ($\approx\!0.014$; four-skill probe per suite, $6$ pairs
each) across all four suites (the bar is Goal's measured range over all $90$
ordered pairs of the ten Goal skills, whose mean is $0.011$); skills are
near-orthogonal everywhere. \emph{Right:} each point is one \emph{negated row}, plotted as its
target-skill SR ($x$) vs.\ mean control-skill SR ($y$), \emph{faceted by suite};
dotted lines mark the taxonomy thresholds (target\,$\leq\!20\%$,
control\,$\geq\!40\%$) and the teal wash is the \emph{selective} quadrant. Goal
populates the selective quadrant with five rows; \textbf{Spatial has none}: its
$t_0$ falls in the collapse corner (target and control both $0\%$) while
$t_1,t_3$ sit in the target-not-suppressed region, so its $\SpatialOff$ aggregate
off-diagonal is an artifact of resistant rows, not locality. Object and Long10
rows sit in the collapse/resistant regions. Marker shape and color both encode
the row class (no red/green-only coding).}
\label{fig:geom_selectivity}
\end{figure*}

\begin{table}[htbp]
\centering
\caption{\textbf{Depth localization} (target SR\,\%, 20 rollouts/cell,
$\alpha{=}1.0$). Negating $\tauv_i$ restricted to one depth band of the
$24$-layer LM stack leaves the skill intact ($80$--$100\%$); only the
full-depth task vector suppresses it ($0\%$). No tested single band suffices at its
native norm; norm-matched band ablations (future work) are needed to conclude
that skill identity is genuinely distributed across depth.}
\label{tab:layerband}
\small
\setlength{\tabcolsep}{6pt}
\begin{tabular}{lcccc}
\toprule
$-\tauv_i$ (diag) & early & mid & late & full ($0$--$23$) \\
\midrule
$t_0$ & 95 & 100 & 90 & \textbf{0} \\
$t_2$ & 80 & 90  & 95 & \textbf{0} \\
$t_5$ & 95 & 90  & 85 & \textbf{0} \\
\bottomrule
\end{tabular}
\end{table}

\begin{table}[htbp]
\centering
\caption{Component ablation for task-0 negation ($\alpha=1.0$).
Target SR $\downarrow$ and mean control SR $\uparrow$ are better. Mean control SR
here is the three-control development panel ($t_1$--$t_3$), the same panel used for
$\alpha^\star$ selection; it is not the nine-control aggregate of the full matrix
(see the denominator ledger), so these $38\%/70\%$ values are not directly
comparable to nine-control means.}
\label{tab:ablation}
\small
\begin{tabular}{lrr}
\toprule
Components negated & Target SR $\downarrow$ & Mean ctrl SR $\uparrow$ \\
\midrule
Full (VLM + AH + PP)  & \textbf{0\%}  & 38\% \\
VLM only              & \textbf{0\%}  & \textbf{70\%} \\
AH only               & 80\%          & 90\% \\
\midrule
None (Goal baseline)  & 100\%         & 100\% \\
\bottomrule
\end{tabular}
\end{table}

\begin{table*}[t]
\centering
\caption{\textbf{Where task-vector negation sits among unlearning approaches}
for VLA skill removal. ``Removal-time data/optimization'' concern the \emph{edit}
at application time only; the ``Upstream artifact'' column makes the hidden cost
explicit (task-vector/merging methods, ours included, need a per-skill expert bank
pre-trained once). Fairness discussion is in the adjacent text.}
\label{tab:comparison}
\small
\setlength{\tabcolsep}{5pt}
\begin{tabular}{lccccl}
\toprule
Method & Removal-time data & Removal-time optimization & VLA & Instant & Upstream artifact \\
\midrule
SISA~\cite{bourtoule2021sisa}              & needs retain set & sharded retraining        & \xmark & \xmark & sharded retrain \\
NegGrad~\cite{graves2021amnesiac}          & needs forget set & gradient steps            & \xmark & \xmark & none \\
Zero-Shot UL~\cite{chundawat2023zero}      & data-free        & gradient steps            & \xmark & \xmark & none \\
VLA-Forget~\cite{ranjan2026vlaf}           & needs forget/retain & module-selective optimization  & \cmark & \xmark & none \\
Task Arith.~\cite{ilharco2023editing}      & data-free        & none (closed-form)        & \xmark & \cmark & expert bank \\
NegMerge~\cite{kim2025negmerge}            & needs merge set  & none (closed-form)        & \xmark & \cmark & expert bank \\
\midrule
Negation (this study)                      & data-free        & none (closed-form)        & \cmark & \cmark & expert bank \\
\bottomrule
\end{tabular}
\end{table*}

\begin{table*}[t]
\centering
\caption{\textbf{Matched-panel gradient-baseline comparison (E1), full detail.} Single target $t_0$; the identical nine-control panel $t_1$--$t_9$; the identical held-out initial states (offset $20$, disjoint from the $0$--$19$ used for $\alpha^\star$ selection); $n{=}20$/cell; one training seed. Gradient baselines at their step-$2000$ (converged) checkpoint. \emph{(A)} Per-control closed-loop successes ($x/20$). \emph{(B)} Per-method aggregates: target $x/20$, raw mean control SR, baseline-normalized retention (vs.\ no-edit), worst control, and removal-time cost. All five removal methods suppress the target ($0/20$); retain-aware optimization does \emph{not} uniformly beat the data-free edit (Gradient-Difference exceeds negation, NegGrad+ does not and zeroes control $t_9$). Normalized retention is the uncapped mean of per-control ratios (edited SR $/$ no-edit SR); the retrain oracle's $102.5\%$ exceeds $100\%$ only because, on a few controls, the leave-one-out retrain scores above the no-edit baseline within $n{=}20$ rollout noise; it is not retention beyond full capability. The pooled intervals reported for these cells are \emph{descriptive} over the nine heterogeneous controls; because the controls are fixed across conditions, each negation-vs-baseline gap is a paired difference over the same initial states. Main Table~\ref{tab:e1} summarizes Panel~B.}
\label{tab:ga}
\small
\setlength{\tabcolsep}{5pt}
\textbf{(A) Per-control successes ($x/20$).}\\[2pt]
\begin{tabular*}{\textwidth}{@{\extracolsep{\fill}}l ccccccccc@{}}
\toprule
Method & $t_1$ & $t_2$ & $t_3$ & $t_4$ & $t_5$ & $t_6$ & $t_7$ & $t_8$ & $t_9$ \\
\midrule
No-edit & 20 & 17 & 17 & 20 & 19 & 18 & 20 & 18 & 19 \\
Negation & 20 & 12 & 7 & 19 & 18 & 9 & 20 & 20 & 8 \\
NegGrad & 0 & 0 & 0 & 0 & 0 & 0 & 0 & 0 & 0 \\
NegGrad+ & 20 & 13 & 13 & 20 & 12 & 17 & 4 & 20 & 0 \\
Grad-Diff & 20 & 16 & 7 & 19 & 19 & 20 & 20 & 20 & 18 \\
Retrain oracle & 20 & 18 & 18 & 20 & 20 & 17 & 20 & 19 & 20 \\
\bottomrule
\end{tabular*}
\\[6pt]
\textbf{(B) Per-method aggregates.}\\[2pt]
\begin{tabular*}{\textwidth}{@{\extracolsep{\fill}}l c c c c l@{}}
\toprule
Method & Target $x/20$ & Raw mean ctrl & Norm.\ retention & Worst ctrl & Removal-time cost \\
\midrule
No-edit & $20/20$ & $93.3\%$ & $100.0\%$ & $85\%$ & --- \\
\textbf{Negation} & $0/20$ & $73.9\%$ & $78.3\%$ & $35\%$ & none (edit-time) \\
NegGrad & $0/20$ & $0.0\%$ & $0.0\%$ & $0\%$ & forget data, grad steps \\
NegGrad+ & $0/20$ & $66.1\%$ & $71.3\%$ & $0\%$ & forget+retain, grad steps \\
Grad-Diff & $0/20$ & $88.3\%$ & $94.1\%$ & $35\%$ & forget+retain, grad steps \\
Retrain oracle & $0/20$ & $95.6\%$ & $102.5\%$ & $85\%$ & full retrain \\
\bottomrule
\end{tabular*}
\end{table*}

\begin{table}[htbp]
\centering
\caption{\textbf{Relearning probe: faster recovery is consistent with behavioral
masking (single $t_0$, one seed).} Closed-loop SR (\%,
20 trials, same fixed initial states across checkpoints) on the suppressed
skill ($t_0$) after $k$ fine-tuning steps, starting from the negated, base
(never-trained-on-$t_0$ task-naive base floor), and goal (intact ceiling)
checkpoints. The diagnostic is the recovery \emph{slope}: at $k{=}250$ the
negated model is already at $85\%$ while the task-naive base floor is still at
$0\%$. This single-skill, single-seed probe is \emph{consistent with} behavioral
masking (the substrate is quickly recoverable); it does not by itself rule out all
forms of weight-level forgetting.}
\label{tab:relearn}
\small
\setlength{\tabcolsep}{3.5pt}
\begin{tabular*}{\columnwidth}{@{\extracolsep{\fill}}lccccc@{}}
\toprule
Checkpoint & $k{=}0$ & $250$ & $500$ & $1000$ & $2000$ \\
\midrule
Base $\thbase$ (task-naive floor)  & 0 & \textbf{0}  & 100 & 95  & 80 \\
Negated $\thedit$               & 0 & \textbf{85} & 95  & 95  & 95 \\
Goal $\thgoal$ (intact ceiling) & 100 & 100 & 85 & 100 & 95 \\
\bottomrule
\end{tabular*}
\\[2pt]
{\footnotesize Negated is observed at $0/20$ (0\%) before relearning.
\textbf{At $k{=}250$}: negated $17/20$ ($85\%$, Wilson $[64,95]\%$) vs.\ scratch
$0/20$ ($0\%$, Wilson $[0,16]\%$), non-overlapping intervals. Base is a noisy
floor thereafter (a full-capability VLA relearns one LIBERO skill within a few
hundred steps once it catches); the intact-ceiling dip to $17/20$ ($85\%$,
Wilson $[64,95]\%$) at $k{=}500$ overlaps $100\%$. Because all checkpoints use
the same fixed initial states, the negated-vs-base gap is a paired difference.
20 rollouts/cell.}
\end{table}
% R6 Gate 5.1: FloatBarrier removed here so Section E text flows onto the same page
% as the Section-D relearn table instead of stranding Table S28 on an isolated,
% vertically-centred float page. (Section-D floats have already placed by this point.)

\section{E. Alignment-Score Diagnostic and Calibration}
\label{asec:alignment}
We recap the score's construction (Supp.\ Sec.~A): a small probe step $\thgoal-\epsilon\tauv_i$ perturbs the goal model's hidden states, and $\mathrm{AS}_i$ is the ratio of the shift induced on skill-$i$ inputs to the mean shift on the other skills' inputs. A skill should be suppressible (here meaning \emph{target-suppression susceptibility}, i.e.\ whether negating $\tauv_i$ drives its own diagonal SR to zero) when $\tauv_i$ moves its own representation more than its neighbors' ($\mathrm{AS}_i\!\gtrsim\!1$); crucially the probe is forward-only and never observes a rollout. We stress that $\mathrm{AS}_i$ ranks \emph{target-suppression susceptibility only}: it says nothing about whether the controls are preserved, so it is not a predictor of selectivity or control retention. Table~\ref{tab:appendix_align} shows this ranking is not a fragile artifact of the probe scale: across $\epsilon\in[0.1,0.5]$ the non-suppressible $t_1$ stays the lowest value ($0.82$--$0.83$) while the suppressible probe skills hold $\geq0.96$. Table~\ref{tab:alignment_law} gives the scale-averaged score for all ten skills. The score \emph{ranks} susceptibility but is not a hard threshold: $t_1,t_4$ score lowest ($0.86$) and are non-suppressible, yet $t_7$ scores above one ($1.03$) while still resisting suppression, so we read $\mathrm{AS}_i$ as a within-suite ranking, not an absolute cutoff.

We note two caveats. Figure~\ref{fig:alignment} (right) probes whether augmenting $\tauv_1$ raises its own-representation shift: every augmentation does, and the $\tauv_1{+}\tauv_3$ direction that empirically flips $t_1$ to suppressible is not uniquely largest, so the ratio does not by itself forecast which augmentation works. More consequentially, Figure~\ref{fig:alignment_calibration} shows $\mathrm{AS}_i$ ranks suppressibility \emph{within} a suite (Spearman $0.75$/AUC $0.87$ on eight Goal skills, tightening to $\rho\!=\!0.72$, $p\!=\!0.019$, AUC $0.90$ at ten), but its absolute threshold does not transfer: on LIBERO-Spatial the ``higher-AS $\Rightarrow$ suppressible'' rule inverts, the sole suppressible skill scoring lowest. We report $\mathrm{AS}_i$ as a within-suite ranking diagnostic, not a portable cutoff.

\begin{table}[htbp]
\centering
\small
\setlength{\tabcolsep}{6pt}
\begin{tabular}{lcccc}
\toprule
$\epsilon$ & $t_0$ & $t_1$ & $t_2$ & $t_3$\\
\midrule
0.1 & 1.21 & \textbf{0.82} & 0.96 & 1.06\\
0.2 & 1.19 & \textbf{0.82} & 0.96 & 1.07\\
0.3 & 1.17 & \textbf{0.82} & 0.97 & 1.08\\
0.4 & 1.16 & \textbf{0.82} & 0.97 & 1.09\\
0.5 & 1.16 & \textbf{0.83} & 0.97 & 1.11\\
\bottomrule
\end{tabular}
\caption{Alignment Score $\mathrm{AS}_i$ per probe scale $\epsilon$
(four-skill probe, $t_0$--$t_3$). The
non-suppressible $t_1$ (bold) scores lowest at every scale, clearly separated
below the suppressible probe skills ($\geq0.96$). The $\mathrm{AS}{=}1$ line is
not a clean separator: suppressible $t_2$ ($0.96$) also sits just below it, so
$\mathrm{AS}_i$ ranks within-suite rather than thresholding at one.}
\label{tab:appendix_align}
\end{table}

\begin{table}[htbp]
\centering
\caption{\textbf{Goal-calibrated target-suppression susceptibility score
$\mathrm{AS}_i$} for all ten Goal skills (averaged over $\epsilon\in[0.1,0.5]$,
forward-only). $\mathrm{AS}_i$ \emph{ranks} within-suite target-suppressibility
(Spearman $0.72$, AUC $0.90$, $n{=}10$) but is not a clean threshold: $t_1,t_4$
(non-suppressible) score lowest ($0.86$), yet $t_7$ scores $1.03$ despite
resisting suppression ($^\dagger$; its diagonal stays at $100\%$). $\mathrm{AS}_i$
does \emph{not} predict control preservation, and its absolute cutoff does not
transfer across suites (Supp.\ Fig.~\ref{fig:alignment_calibration}). ``Suppressible''
$=$ diagonal SR $\leq\!20\%$ under negation (target suppression only). This marks
\emph{seven} skills: the five separation rows plus the two global-collapse rows
$t_8,t_9$, which suppress their own target yet destroy all controls and so are
excluded from the ``five'' target--control \emph{separation} skills discussed in the
main text.}
\label{tab:alignment_law}
\small
\begin{tabular}{llrrc}
\toprule
Skill & Description & $\mathrm{AS}_i$ & Diag SR & Suppressible? \\
\midrule
$t_0$ & open mid drawer      & 1.21 & 0   & \cmark \\
$t_2$ & wine on cabinet      & 1.05 & 0   & \cmark \\
$t_3$ & top drawer+bowl      & 0.97 & 5   & \cmark \\
$t_5$ & push plate           & 1.10 & 0   & \cmark \\
$t_6$ & cream cheese in bowl & 1.03 & 15  & \cmark \\
$t_8$ & bowl on plate        & 1.01 & 0   & \cmark \\
$t_9$ & wine on rack         & 1.04 & 0   & \cmark \\
\midrule
$t_1$ & bowl on stove        & \textbf{0.86} & 100 & \xmark \\
$t_4$ & bowl on cabinet      & \textbf{0.86} & 30  & \xmark \\
$t_7$ & turn on stove        & 1.03 & 100 & \xmark\,$^\dagger$ \\
\bottomrule
\end{tabular}
\end{table}

\begin{figure*}[htbp]
\centering
\includegraphics[width=\textwidth]{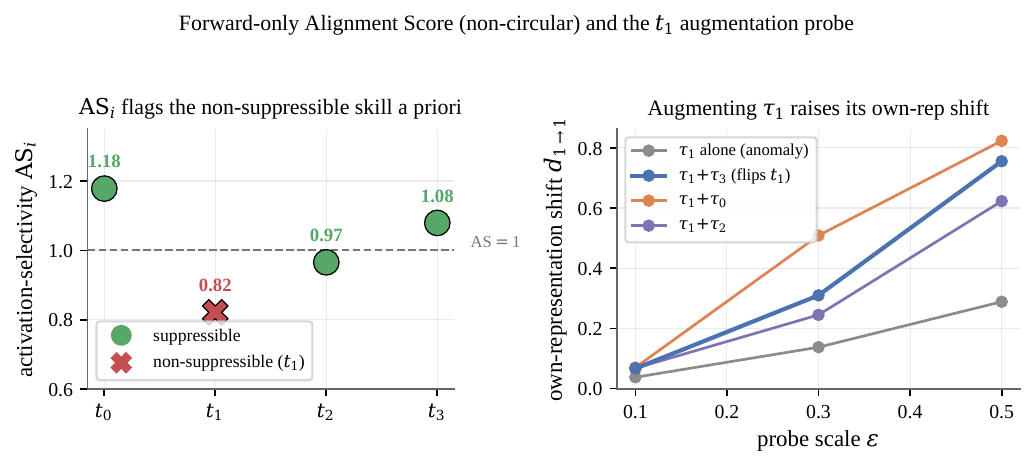}
\caption{\textbf{Alignment Score $\mathrm{AS}_i$, and a $t_1$
augmentation probe.} \emph{Left:} Alignment Score $\mathrm{AS}_i$ on the
four-skill probe ($t_0$--$t_3$) vs.\ ground-truth suppressibility. The
non-suppressible $t_1$ (\xmark, $\mathrm{AS}{=}0.82$) scores lowest, well below
the suppressible skills; the $\mathrm{AS}{=}1$ line is not a hard separator
(suppressible $t_2$ sits just under it at $0.96$), consistent with the
within-suite ranking reading. Scores here are relative to the three other probe
skills; the ten-skill headline values (Table~\ref{tab:alignment_law}, $t_1{=}0.86$)
use all nine others. No closed-loop outcome enters the score.
\emph{Right:} the own-representation
shift $d_{1\to1}$ when $\tauv_1$ is augmented by a second skill vector.
Every augmentation raises $t_1$'s own-shift above $\tauv_1$-alone; the
$\tauv_1{+}\tauv_3$ direction (which we \emph{measured} flips $t_1$ to
suppressible, Experiments) is among the largest, but so is
$\tauv_1{+}\tauv_0$, so own-shift magnitude, not the selectivity ratio,
is what an augmentation raises. The ratio does not
by itself forecast which augmentation flips $t_1$.}
\label{fig:alignment}
\end{figure*}

\begin{figure*}[htbp]
\centering
\includegraphics[width=\textwidth]{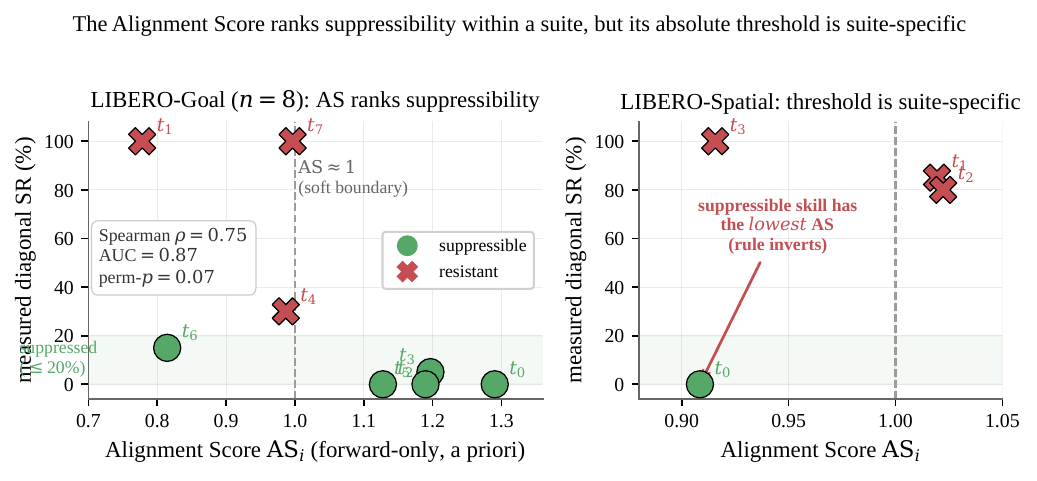}
\caption{\textbf{The Alignment Score ranks suppressibility within a suite, but
its absolute threshold is suite-specific.}
\emph{Left:} on LIBERO-Goal ($n{=}8$), $\mathrm{AS}_i$ (forward-only,
a~priori) vs.\ measured diagonal suppression. Suppressible skills (green) score
high and the resistant outliers (red $\times$) score near/at the
$\mathrm{AS}\!\approx\!1$ boundary (Spearman $0.75$, AUC $0.87$ on these
eight skills; extending to ten skills tightens this to $\rho\!=\!0.72$,
$p\!=\!0.019$, AUC $0.90$, the Alignment Score).
\emph{Right:} the same score on LIBERO-Spatial, where only $t_0$ is
suppressible: the Goal-calibrated ``higher-AS $\Rightarrow$ suppressible'' rule
\emph{inverts} (the suppressible skill has the lowest AS). We report this
negative: $\mathrm{AS}_i$ is a within-suite ranking diagnostic,
not a portable threshold.}
\label{fig:alignment_calibration}
\end{figure*}

\section{F. Flow-Matching Backbones ($\pi_{0.5}$)}
Finally, we replicate the full analysis on the flow-matching \piZeroFive{} backbone (3B), whose continuous velocity-field action head differs from both MergeVLA's regression head and OpenVLA's discrete token decoder; confirming the effect here establishes it across a third action-head class rather than as an artifact of one decoding scheme.

\begin{table}[htbp]
\centering
\small
\setlength{\tabcolsep}{5pt}
\begin{tabular}{lcccc}
\toprule
Negated & \multicolumn{4}{c}{Eval skill (SR \%, $n\!=\!20$)} \\
\cmidrule(lr){2-5}
expert & $s_0$ & $s_1$ & $s_2$ & $s_3$ \\
\midrule
$-\tauv_{0}$ & \underline{\textbf{0}} & 90 & 70 & 30 \\
$-\tauv_{1}$ & 90 & \underline{\textbf{0}} & 50 & 65 \\
$-\tauv_{2}$ & 65 & 35 & \underline{\textbf{0}} & 70 \\
$-\tauv_{3}$ & 80 & 0 & 10 & \underline{\textbf{0}} \\
\midrule
baseline (no edit) & 100 & 100 & 95 & 100 \\
\bottomrule
\end{tabular}
\caption{\textbf{\piZeroFive{} (3B, flow-matching) per-skill negation matrix
on LIBERO-\textsc{Goal}} ($n\!=\!20$/cell; also the third panel of
Fig.~\ref{fig:dualheat}). Skills $s_0$--$s_3$ are four distinct
\texttt{libero\_goal} tasks ($s_0{=}t_7,s_1{=}t_8,s_2{=}t_1,s_3{=}t_2$). Row
$i$: negate expert $\tauv_i$ on the multitask policy. All four targets are
\emph{suppressed}: every diagonal (\underline{\textbf{target}}) is driven to
$0\%$ (mean $0\%$, Wilson $[0,5]\%$) from a $95$--$100\%$ baseline, while
off-diagonal \emph{controls} are retained at $54.6\%$ (pooled-descriptive Wilson
$[48,61]\%$ over the $12$ off-diagonal cells; per-row control means
$63.3/68.3/56.7/30.0\%$), a $54.6$\,pp aggregate gap comparable to the continuous
MergeVLA (nine-control $53.3$\,pp) and discrete OpenVLA ($\sim\!51$\,pp) backbones. All four
per-row target-vs-mean-control gaps are positive ($30.0$--$68.3$\,pp); under the
taxonomy threshold (target $\leq\!20\%$, mean control $\geq\!40\%$) three of the
four rows ($s_0,s_1,s_2$) show target--control separation, while row $s_3$ ($=t_2$)
suppresses its target ($0\%$) but retains only $30\%$ mean control (a positive-gap
but non-selective row). Thus all four targets are suppressed, all four rows are
positive-gap, and three of four rows are selective.}
\label{tab:pi05goalmatrix}
\end{table}

Table~\ref{tab:pi05matrix} gives the per-skill negation matrix on LIBERO-\textsc{Long}. Every diagonal target is driven to $0\%$, but the off-diagonal controls also collapse (mean $3.3\%$ from an $80$--$100\%$ baseline), leaving only a $3.3$\,pp selectivity gap versus the $54.6$\,pp gap on the same backbone's LIBERO-\textsc{Goal} matrix. Despite $n\!=\!10$ the pattern is descriptively clear: the worst-case control (Wilson $[6,51]\%$) does not overlap its no-edit baseline (Wilson $[72,100]\%$). This reproduces the Object/Long10 selectivity breakdown on a flow-matching expert bank, consistent with the boundary being behavioral (densely shared long-horizon primitives) and cross-architectural, not tied to the continuous/discrete backbones.

Where skills are separable, selectivity returns. The whole-model dose--response on LIBERO-Object task~0 (Table~\ref{tab:pi05dose}) falls monotonically from $85\%$ to $0\%$ as $\alpha$ rises from $0$ to $1.0$ (Spearman $\rho\!=\!-1.0$, exact one-sided permutation $p\!=\!1/5!\!\approx\!0.008$ over the five $\alpha$ points; baseline-vs-negation two-sided Fisher exact $p\!=\!2.6\!\times\!10^{-8}$ at $n\!=\!20$), with a second task replicating ($100\%\!\to\!10\%$). The per-skill LIBERO-\textsc{Goal} curve (Table~\ref{tab:pi05goaldose}) collapses the target to $0\%$ once $\alpha\!\ge\!0.75$ while the control holds at $90$--$100\%$ throughout, so selectivity spans a broad operating band rather than a knife-edge $\alpha$.

The asymmetry then closes: additive composition $\theta_{\mathrm{ft}}+\sum_i\tauv_i$ (Table~\ref{tab:pi05compose}) drives \emph{every} evaluated skill to $0\%$ on \emph{both} suites (including the Goal skills that negation selectively suppresses), matching the continuous and discrete backbones. We caveat that the flow-matching cells use $n\!=\!10$ rollouts, coarser than the $n\!=\!20$ used elsewhere.

\begin{table}[htbp]
\centering
\small
\setlength{\tabcolsep}{5pt}
\begin{tabular}{lcccc}
\toprule
Negated & \multicolumn{4}{c}{Eval skill (SR \%, $n\!=\!10$)} \\
\cmidrule(lr){2-5}
expert & $s_0$ & $s_1$ & $s_2$ & $s_3$ \\
\midrule
$-\tauv_{0}$ & \underline{\textbf{0}} & 0 & 0 & 0 \\
$-\tauv_{1}$ & 0 & \underline{\textbf{0}} & 0 & 0 \\
$-\tauv_{2}$ & 0 & 10 & \underline{\textbf{0}} & 10 \\
$-\tauv_{3}$ & 0 & 20 & 0 & \underline{\textbf{0}} \\
\midrule
baseline (no edit) & 80 & 100 & 100 & 90 \\
\bottomrule
\end{tabular}
\caption{\textbf{\piZeroFive{} (3B, flow-matching) per-skill negation matrix
on LIBERO-\textsc{Long}} ($n\!=\!10$/cell). Skills $s_0$--$s_3$ are four
distinct \texttt{libero\_10} long-horizon tasks (e.g.\ ``put both the
alphabet soup and the cream cheese box in the basket''). Row $i$: negate
expert task vector $\tauv_i$ on the multitask policy; columns: closed-loop SR
on each skill. Diagonal (\underline{\textbf{target}}) is suppressed everywhere ($0\%$),
but off-diagonal \emph{controls} also collapse (mean $3.3\%$), a
$3.3$\,pp selectivity gap vs.\ $54.6$\,pp on the same backbone's Goal matrix. The pattern is
unambiguous despite $n\!=\!10$: every control drops from an $80$--$100\%$
baseline to $\leq20\%$, and even the worst-case control (Wilson
$[6,51]\%$) does not overlap its own no-edit baseline (Wilson $[72,100]\%$),
so the collapse is outside the per-cell noise band. This
\emph{reproduces} the Object/Long10 selectivity breakdown
(\S\,Generality and Boundary) on a flow-matching expert bank: the boundary is
behavioral (densely shared long-horizon primitives) and cross-architectural,
not specific to the continuous or discrete-token action heads.}
\label{tab:pi05matrix}
\end{table}

\begin{table}[htbp]
\centering
\small
\setlength{\tabcolsep}{4.5pt}
\begin{tabular}{lccccc}
\toprule
& \multicolumn{5}{c}{Negation strength $\alpha$} \\
\cmidrule(lr){2-6}
Task~0 SR (\%) & $0$ & $0.25$ & $0.5$ & $0.75$ & $1.0$ \\
\midrule
point est. & $85$ & $70$ & $30$ & $15$ & $0$ \\
95\% Wilson & \tiny$[64,95]$ & \tiny$[48,85]$ & \tiny$[15,52]$ & \tiny$[5,36]$ & \tiny$[0,16]$ \\
\bottomrule
\end{tabular}
\caption{\textbf{\piZeroFive{} (3B, flow-matching) closed-loop negation
dose--response} on LIBERO-Object task~0, $n\!=\!20$ trials/cell. Success
rate of the negated policy $\theta_{\mathrm{LIBERO}}-\alpha\tauv$ collapses
monotonically as the LIBERO task vector is subtracted (Spearman
$\rho\!=\!-1.0$, exact one-sided permutation $p\!\approx\!0.008$ over five $\alpha$ points; the subtracted $\tauv$ is the whole-suite LIBERO-Object task vector, not a per-skill vector). Baseline vs.\ full negation is cleanly
separated (Fisher $p\!<\!10^{-7}$; non-overlapping Wilson intervals), and a
second task replicates ($100\%\!\to\!10\%$). This is closed-loop target suppression on
the tested \piZeroFive{} policy, complementing the weight-space
localization and the SmolVLA~\cite{shukor2025smolvla} action-fidelity result (Supp.\ Sec.~H).}
\label{tab:pi05dose}
\end{table}

\begin{table}[htbp]
\centering
\small
\setlength{\tabcolsep}{5pt}
\begin{tabular}{lccccc}
\toprule
$\pi_{0.5}$-Goal $s_0$ & \multicolumn{5}{c}{Negation strength $\alpha$} \\
\cmidrule(lr){2-6}
SR (\%, $n\!=\!10$) & $0.25$ & $0.5$ & $0.75$ & $1.0$ & $1.25$ \\
\midrule
target ($t_7$)   & $100$ & $100$ & \textbf{0} & \textbf{0} & \textbf{0} \\
control ($t_8$)  & $100$ & $100$ & $100$ & $90$ & $90$ \\
\bottomrule
\end{tabular}
\caption{\textbf{Per-skill negation dose--response on the flow-matching
\piZeroFive{} (LIBERO-\textsc{Goal} $s_0$)}, $n\!=\!10$/cell. The target
collapses to $0\%$ once $\alpha\!\ge\!0.75$ while the control is retained at
$90$--$100\%$ throughout: selectivity holds across a broad $\alpha$ band, not
only at $\alpha\!=\!1$.}
\label{tab:pi05goaldose}
\end{table}

\begin{table}[htbp]
\centering
\small
\setlength{\tabcolsep}{5pt}
\begin{tabular}{lcccc}
\toprule
\piZeroFive{} & \multicolumn{4}{c}{Eval skill (closed-loop SR \%, $n\!=\!10$)} \\
\midrule
LIBERO-\textsc{Goal} & $s_0$ & $s_1$ & $s_2$ & $s_3$ \\
\ \ multitask baseline & 100 & 100 & 100 & 100 \\
\ \ $+\sum_i\tauv_i$ (compose) & \textbf{0} & \textbf{0} & \textbf{0} & \textbf{0} \\
\midrule
LIBERO-\textsc{Long} & $s_0$ & $s_1$ & $s_2$ & $s_3$ \\
\ \ multitask baseline & \multicolumn{4}{c}{$80$--$100$} \\
\ \ $+\sum_i\tauv_i$ (compose) & \textbf{0} & \textbf{0} & \textbf{0} & \textbf{0} \\
\bottomrule
\end{tabular}
\caption{\textbf{Additive composition collapses the flow-matching
\piZeroFive{} policy on every skill} ($n\!=\!10$/cell). Adding the four
per-skill task vectors to the multitask policy
($\theta_{\mathrm{ft}}+\sum_i\tauv_i$) drops closed-loop success to $0\%$ on
all four evaluated skills of \emph{both} suites, including LIBERO-\textsc{Goal},
where \emph{negating} the same vectors is selective
(main paper). This completes the
negation-vs-composition asymmetry on the flow-matching backbone, matching what we
observe on the continuous-regression and discrete-token action heads.}
\label{tab:pi05compose}
\end{table}

\section{G. Diagnostics: Sharpness, Path, Task-1, and Alignment (extended)}
\label{asec:diagextended}
\label{sec:mechanism}

Having shown the asymmetry across suites and backbones, we now ask what observable properties co-vary with it and help characterize it.

\subsection{$\varepsilon$-Sharpness and the Composition/Negation Asymmetry}

Because raw $\varepsilon$-sharpness is parameterization-dependent~\cite{dinh2017sharp}, we report it under adaptive (ASAM~\cite{kwon2021asam}) and filter-normalized~\cite{li2018visualizing} metrics (Supp.\ Table~\ref{tab:sharpness}, Supp.\ Fig.~\ref{fig:sharpness}): single-task experts are extremely sharp ($14$--$18\times$), the multitask goal model moderately sharp ($4$--$7\times$), and negated models flattest ($\sim$2.4$\times$), consistent with negation moving into a flatter region while composition moves toward a higher-loss region between the sharp expert basins (an observed closed-loop collapse along that path, Fig.~\ref{fig:barrier})~\cite{foret2021sam,garipov2018mode,frankle2020lmc}. This is a \emph{landscape signature} that co-varies with the suppressed state (\emph{which} skills are suppressible is given by alignment, Supp.\ Sec.~E). The ordering survives three confounds: absolute loss increase (Expert $285\gg$ Goal $92\gg$ Negated $58$), matched weight norm ($\|\theta\|\approx 1937$), and scale-normalized (ASAM/filter-norm) metrics (Supp.\ Table~\ref{tab:sharpness_v2}), so we treat it as co-varying with the asymmetry. A flat-vs-sharp intervention associates flatness with \emph{selectivity}: we re-train a task-$0$ expert with SAM~\cite{foret2021sam} so it occupies a flatter basin ($100\%$ own-task SR, matched to the default), then negate it. At matched perturbation norm ($\alpha\!=\!1.95$, $\lVert\Delta\theta\rVert\!=\!84$ vs.\ the default's $83$), the flat expert's negation \emph{does} suppress the target ($0\%$), so flatness does not prevent suppression, but it \emph{also} collapses the control ($t_2$: $30\%$ for the sharp expert $\to\!0\%$ for the flat one), whereas the sharp expert suppresses selectively. This single-expert intervention associates flatness with the \emph{selectivity} of suppression rather than with whether suppression occurs; we note it also changes the task-vector direction and training trajectory, so we read it as intervention-based association, not an isolated causal test. Sharpness and alignment are complementary: across the four negated models we see no monotone sharpness--suppressibility trend ($\rho\!=\!0.4$, $p\!=\!0.6$, $n\!=\!4$; at $n{=}4$ the correlation CI spans essentially $[-1,1]$, so we read this only as the absence of a strong trend).

\paragraph{Closed-loop path behavior along the two edit directions.}
We interpolate along both edit directions from $\thgoal$ and measure closed-loop success (Fig.~\ref{fig:barrier}): negation $\theta(\beta)=\thgoal-\beta\tauv_0$ and composition $\theta(\beta)=\thgoal+\beta(\tauv_0{+}\tauv_1)$, with $\beta$ playing the same magnitude role as $\alpha$ in Eq.~\ref{eq:negation}, $\beta\in[0,1.5]$ (20 rollouts/point; VLM weights only). Negation descends into a selective plateau (target $\to\!0\%$ at $\beta{=}1.0$, control retained at $90\%$), whereas composition never has such a point: the control collapses \emph{before} the target ($\beta{=}1.0$: target $65\%$, control $0\%$). This barrier is \emph{behavioral}, not a static-loss artifact: along the composition path the teacher-forced loss rises \emph{less} than along negation's ($+0.05$ vs.\ $+0.09$ at $\beta{=}1$), yet only composition collapses the policy. Across a five-pair population (Supp.\ Table~\ref{tab:barrier_pairs}) composition never reaches a selective point in any pair, whereas negation reaches one in $2/5$ pairs whose skills are behaviorally separable (cleanest $t_5/t_4$: target $\to\!0\%$, control $100\%$; also $t_0/t_2$) and fails to isolate in the more entangled pairs ($t_3/t_0$, $t_5/t_6$, $t_2/t_5$), the same separability boundary as \S\,Generality and Boundary. Along the tested goal-anchored positive path, closed-loop controls thus fail before the target is suppressed, despite a smaller teacher-forced loss increase, reminiscent of (though not a formal instance of) mode-connectivity loss barriers~\cite{garipov2018mode,frankle2020lmc}. The figure summarizing this path (Fig.~\ref{fig:barrier}) is promoted to the main-paper Discussion.

\subsection{Task-1 Anomaly: Direction Misalignment}
\label{sec:task1}

Negating $t_1$ has \emph{zero target effect} (Supp.\ Table~\ref{tab:effect}, $e_{11}=0.00$) despite $\|\tauv_{t_1}^W\|=78.9$, comparable to $\|\tauv_{t_0}^W\|=83.4$ and $\|\tauv_{t_2}^W\|=89.7$; skill vectors are near-orthogonal ($\overline{\cos}(\tauv_{t_i}, \tauv_{t_j}) \approx 0.014$ on the four-skill probe, $0.011$ over all ten Goal skills) and VLM-only negation also yields $\mathrm{SR}_{t_1}=100\%$, so the tested cosine geometry and the AH-component ablation do not explain the anomaly. These results are \emph{consistent with} $\tauv_{t_1}$'s VLM direction being misaligned with the goal model's internal encoding of task~1; they do not rule out all forms of cross-task overlap. Supporting the misalignment reading, simultaneous negation of $\{t_1, t_3\}$ achieves $\mathrm{SR}_{t_1}=0\%$.

\subsection{A Within-Suite Target-Suppression Diagnostic}
\label{asec:targdiag}
\label{sec:alignment_law}

Can we predict, \emph{before} negating and using only the goal model and $\tauv_i$, whether a skill's \emph{target} will be suppressed? We use the \textbf{Alignment Score} $\mathrm{AS}_i$ (Eq.~\ref{eq:alignment_score}): take a probe step $\thgoal - \epsilon\,\tauv_i$ ($\epsilon \ll \alpha$) and measure how much it shifts the goal model's hidden states on skill-$i$ inputs relative to other skills':
\begin{equation}
\mathrm{AS}_i = \frac{\mathbb{E}_{x \sim \mathcal{D}_i}\,\lVert h(\thedit;x) - h(\thgoal;x)\rVert}
{\frac{1}{N-1}\sum_{k\neq i}\mathbb{E}_{x \sim \mathcal{D}_k}\,\lVert h(\thedit;x) - h(\thgoal;x)\rVert}.
\label{eq:alignment_score}
\end{equation}
A skill's target tends to be suppressible when $\tauv_i$ perturbs its own representations more than others' ($\mathrm{AS}_i \gtrsim 1$). The score is non-circular (forward-only) and cheap. We stress it is a within-suite \emph{target-suppression} susceptibility diagnostic: it ranks which targets can be driven down, and does \emph{not} predict control preservation (selectivity); indeed $t_8,t_9$ score as suppressible yet cause non-selective global collapse.

Supp.\ Table~\ref{tab:appendix_align} shows $\mathrm{AS}_i$ separates non-suppressible task~1 ($\mathrm{AS}=0.82$) from every suppressible task ($\mathrm{AS}\geq0.96$) across every probe scale; we read it for ranking and margin, not an absolute cutoff. On the $8$-task ground truth (the eight non-collapse rows of Fig.~\ref{fig:goal10main}) it gives Spearman $\rho\!=\!0.75$ and ROC-AUC $0.87$; extending to $n{=}10$ ($t_8,t_9$) gives $\rho\!=\!0.72$ and ROC-AUC $0.90$, both new skills correctly ranked. We stress the small-$n$ uncertainty: a nonparametric bootstrap ($10$k resamples over the $n{=}8$ tasks) gives wide intervals (Spearman $95\%$ CI $[0.19,0.96]$, AUC CI $[0.50,1.00]$), so this is a suggestive within-suite ranking, not a tight estimate. Against the per-skill suppression threshold $\alpha^{\mathrm{sup}}_i$, $\mathrm{AS}_i$ and $\alpha^{\mathrm{sup}}_i$ are negatively rank-correlated (Spearman $\rho\!=\!-0.78$, CI $[-0.99,-0.17]$; Kendall $\tau\!=\!-0.70$; $n\!=\!8$). Because we report several correlation tests here, we treat individual $p$-values as exploratory and apply Benjamini--Hochberg control across the alignment tests; the graded $\mathrm{AS}$--$\alpha^{\mathrm{sup}}$ association and the $n{=}10$ ranking survive at $q\!<\!0.05$, while the borderline $n{=}8$ Spearman ($p\!=\!0.04$) does not. \textbf{The calibrated threshold does \emph{not} transfer across suites (a negative result):} on LIBERO-Spatial the suppressible skill $t_0$ has the \emph{lowest} score ($\mathrm{AS}\!=\!0.91$, rank $4/4$), inverting the ordering (Supp.\ Fig.~\ref{fig:alignment_calibration}); this is not a similarity artifact (mean pairwise VLM cosine $\approx0.014$ on the four-skill probe for both suites), and within-suite correlation is $0.26$ ($p{=}0.74$) on Object and $0.00$ ($p{=}1.0$) on Long10 (vs.\ Goal's $0.75$). We report $\mathrm{AS}_i$ as calibrated on LIBERO-Goal with a threshold that neither transfers nor reliably holds elsewhere; the absolute threshold does not transfer across suites and the within-suite correlation is weak-to-absent outside Goal, so we report the diagnostic as within-suite only: cross-suite selectivity is not captured by weight-space cosine geometry, consistent with a behavioral-overlap account.

\paragraph{Does the alignment signal transfer to OpenVLA-7B?}
\label{sec:openvla_align}
On OpenVLA-7B the three high-baseline target skills are suppressible (Supp.\ Table~\ref{tab:openvla}; $t_3$ excluded as non-diagnostic), so we test whether the shift term predicts the graded off-diagonal collateral over 12 cells. Probing all unified weights yields no signal ($\rho\!=\!-0.24$; a common-mode term from the shared vision encoder), but matching the MergeVLA protocol, perturbing only the language-model weights (fixed a~priori by the VLM-only ablation, Supp.\ Sec.~D, Table~\ref{tab:ablation}, where VLM-only negation reached $0\%$ target with $70\%$ vs.\ $38\%$ retention on the three-control development panel), restores it: every $\mathrm{AS}_i\!>\!1$ ($1.14$--$1.41$) and the collateral correlation flips positive and scale-stable ($\rho\!=\!+0.30,+0.40,+0.41$ at $\epsilon=0.1,0.3,0.5$). The signal thus \emph{partially} transfers: direction-consistent under matched support but underpowered: with $n\!=\!12$ cells, $\lvert\rho\rvert\!\geq\!0.58$ is needed for $p\!<\!0.05$ and $\rho\!=\!0.40$ reaches significance only at $n\!\gtrsim\!25$. We thus report the alignment score as empirically associated with suppressibility on MergeVLA-Goal and direction-consistent (not-yet-significant) on OpenVLA; what recurs across the tested policies is the negation/composition asymmetry itself (Supp.\ Table~\ref{tab:openvla}), not its quantitative predictor.
\section{H. Localization and Flow-Matching Extensions}
\label{asec:localization}
\label{sec:extensions}

Supporting studies (\emph{what} is edited, \emph{how robustly}, \emph{where else})
use $10$--$20$ rollouts/cell with Wilson intervals; several are individually
underpowered, so we read them together as a \emph{convergent set of evidence}.

\subsection{Component and Geometry Localization}

The task-vector energy and the tested suppression effect concentrate in the VLM
component: VLM-only negation matches full negation on target suppression while
better preserving control skills, and VLM task vectors are near-orthogonal across
skills (Supp.\ Table~\ref{tab:ablation},~Supp.\ Fig.~\ref{fig:geometry}).

\paragraph{Task-vector energy concentrates in the VLM block across backbones; component-restricted suppression is shown on MergeVLA and $\pi_{0.5}$.} Beyond our continuous-L1 (MergeVLA) and
discrete-token (OpenVLA) heads, we test two \emph{flow-matching} backbones:
\textbf{SmolVLA}~\cite{shukor2025smolvla} ($0.45$B) and
\textbf{$\pi_{0.5}$}~\cite{intelligence2025pi05} ($3$B). Here
$\theta_{\mathrm{LIBERO}}$ (equivalently $\theta_{\mathrm{ft}}$ for $\pi_{0.5}$)
denotes the multitask fine-tuned anchor, the flow-matching analogue of
$\thgoal$; the base is $\theta_{\mathrm{base}}$. Forming
$\tauv = \theta_{\mathrm{LIBERO}} - \theta_{\mathrm{base}}$, the task-vector
$\ell_2$ norm concentrates in one block, reported here as the whole-vector norm
ratio $\lVert\tauv_{\mathrm{block}}\rVert/\lVert\tauv\rVert$ (not the squared-energy
share of Fig.~\ref{fig:geometry}): for SmolVLA this ratio is
$98.9\%$ for the vision-language$+$expert block; for
$\pi_{0.5}$ it is $99.9\%$ ($\lVert\tauv\rVert=242$) for its
\texttt{paligemma\_with\_expert} block. The effect \emph{localizes} to this
block: restricting per-skill negation on $\pi_{0.5}$-Goal to the VLM alone still
drives the target to $0\%$, whereas negating only the action-expert leaves it at
$100\%$. As on
MergeVLA (Supp.\ Table~\ref{tab:layerband}), no single depth band suffices: early
($0$--$8$) or late ($18$--$26$) leave the skill at $100\%$ and mid ($9$--$17$)
reaches only $70\%$.

Negation also \emph{suppresses the target in closed loop}. On \piZeroFive{}, the
negated policy collapses monotonically (Supp.\ Table~\ref{tab:pi05dose}, Spearman
$\rho\!=\!-1.0$): from $85\%$ at $\alpha\!=\!0$ through $70\%,\,30\%,\,15\%$ to
$0\%$ at $\alpha\!=\!1$ (Fisher $p\!<\!10^{-7}$), replicating on a second task
($100\%\!\to\!10\%$). Full negation across four suites ($n\!=\!20$, task~0)
collapses three (Goal $95\%\!\to\!0\%$, Object $85\%\!\to\!0\%$, Long10
$60\%\!\to\!0\%$), while only Spatial partially survives ($90\%\!\to\!30\%$). For SmolVLA we do not run a component-restricted closed-loop ablation; the SmolVLA
evidence is the task-vector norm concentration above and an open-loop action-output
signal: negation drives the predicted-action cosine to the demonstrator from
$0.92$ through $0.81$ ($\alpha\!=\!0.5$) to $0.26$ ($\alpha\!=\!1$). Thus MergeVLA
and $\pi_{0.5}$ carry component-restricted suppression evidence, whereas SmolVLA
contributes norm concentration and action-output corroboration, not the same
closed-loop localization.

\paragraph{A flow-matching per-skill negation matrix.}
On \textbf{LIBERO-\textsc{Goal}} negation shows a clear \emph{target--control
separation} (Fig.~\ref{fig:dualheat}, right; full matrix Supp.\ Table~\ref{tab:pi05goalmatrix}): from a $95$--$100\%$ baseline, every diagonal target
drops to $0\%$ (pooled-descriptive Wilson $[0,5]\%$) while off-diagonal controls are
retained at $54.6\%$ (pooled-descriptive Wilson $[48,61]\%$; per-row control means
$63.3/68.3/56.7/30.0\%$): a $54.6$\,pp aggregate gap, all four per-row gaps positive
($30.0$--$68.3$\,pp), with three of the four rows selective under the
target$\leq\!20\%$/control$\geq\!40\%$ threshold (row $s_3$ suppresses its target
but retains only $30\%$ mean control), so per-skill negation is selective on
$3/4$ flow-matching rows and positive-gap on all four ($n\!=\!20$/cell). Sweeping strength (Supp.\ Table~\ref{tab:pi05goaldose}) shows a
robust band: the target holds at $100\%$ through $\alpha\!=\!0.5$, collapses to
$0\%$ for all $\alpha\!\ge\!0.75$, while the control stays at $90$--$100\%$. On
\textbf{LIBERO-\textsc{Long}} the same construction gives a \emph{negative}
result (Supp.\ Table~\ref{tab:pi05matrix}): every target is suppressed ($0\%$, $92.5\%$
baseline) but controls collapse with it (off-diagonal mean $3.3\%$; gap
$3.3$\,pp), reproducing the boundary for the token/L1 backbones. Target--control
separation is thus not confined to one tested architecture, but it remains both
suite-dependent and model-dependent across the backbones we test.

\paragraph{Composition collapses the flow-matching policy too.} The additive
counterpart $\theta_{\mathrm{ft}} + \alpha\sum_i \tauv_i$
($\alpha\!=\!1$; Supp.\ Table~\ref{tab:pi05compose}) drives \emph{every} tested skill
to $0\%$ on both suites: on LIBERO-\textsc{Goal}, where the same vectors
\emph{negate} selectively (targets $\to\!0\%$, controls $54.6\%$ at $n\!=\!20$), and on
LIBERO-\textsc{Long} from an $80$--$100\%$ baseline. Thus subtraction produces target--control separation while goal-anchored
re-addition of the same vectors collapses the policy; this asymmetry is
consistent across all three action-head classes: continuous regression
(MergeVLA), discrete tokens (OpenVLA), and flow matching (\piZeroFive{}).

% Reproducibility Checklist omitted from the arXiv release (kept only in the
% AAAI submission build). Body + appendix render via \mainfalse above.
\fi

\end{document}